%% file: main.tex
\documentclass[10pt,twocolumn,letterpaper]{article}

\usepackage[]{cvpr} 

\input{preamble}

\definecolor{cvprblue}{rgb}{0.21,0.49,0.74}
\usepackage[pagebackref,breaklinks,colorlinks,citecolor=cvprblue]{hyperref}

\usepackage[accsupp]{axessibility} 
\usepackage{multirow}
\usepackage{svg}

\def\paperID{5187} 
\def\confName{CVPR}
\def\confYear{2024}

\title{Generating Enhanced Negatives for Training Language-Based Object Detectors}

\author{
Shiyu Zhao$^{1,*}$ $\quad$ Long Zhao$^{3}$ $\quad$ Vijay Kumar B G$^{2}$ $\quad$ Yumin Suh$^{2}$ \\ Dimitris N. Metaxas$^{1}$ $\quad$ Manmohan Chandraker$^{2,4}$ $\quad$ Samuel Schulter$^{2}$\\
\\
$^{1}$ Rutgers University $\quad$ $^{2}$ NEC Laboratories America $\quad$ $^{3}$ Google Research $\quad$ $^{4}$ UC San Diego
}

\begin{document}
\maketitle
\input{sec/0_abstract}
\blfootnote{$^{*}$ Part of this work was done during an internship at NEC Laboratories America. Correspondence to: Shiyu Zhao \hyperlink{mailto:sz553@rutgers.edu}{sz553@rutgers.edu}}
\vspace{-0.75cm}

\input{sec/1_intro}

\input{sec/2_related_work}

\input{sec/3_method}

\input{sec/4_experiment}

\input{sec/5_conclusion}

{
    \small
    \bibliographystyle{ieeenat_fullname}
    \bibliography{main}
}

\input{sec/X_suppl}


\end{document}

%% file: preamble.tex
\usepackage[dvipsnames]{xcolor}

\makeatletter
\renewcommand{\paragraph}{%
  \@startsection{paragraph}{4}%
  {\z@}{0.5ex \@plus 1ex \@minus .2ex}{-1em}%
  {\normalfont\normalsize\bfseries}%
}
\makeatother

\usepackage{listings}
\usepackage{float}
\newfloat{lstfloat}{htbp}{lop}
\floatname{lstfloat}{Listing}

\newcommand\blfootnote[1]{%
  \begingroup
  \renewcommand\thefootnote{}\footnote{#1}%
  \addtocounter{footnote}{-1}%
  \endgroup
}

%% file: sec/0_abstract.tex
\begin{abstract}

The recent progress in language-based open-vocabulary object detection can be largely attributed to finding better ways of leveraging large-scale data with free-form text annotations. Training such models with a discriminative objective function has proven successful, but requires good positive and negative samples. However, the free-form nature and the open vocabulary of object descriptions make the space of negatives extremely large. Prior works randomly sample negatives or use rule-based techniques to build them. In contrast, we propose to leverage the vast knowledge built into modern generative models to automatically build negatives that are more relevant to the original data. Specifically, we use large-language-models to generate negative text descriptions, and text-to-image diffusion models to also generate corresponding negative images. Our experimental analysis confirms the relevance of the generated negative data, and its use in language-based detectors improves performance on two complex benchmarks. Code is available at \url{https://github.com/xiaofeng94/Gen-Enhanced-Negs}.

\end{abstract}

%% file: sec/1_intro.tex
\section{Introduction}
\label{sec:intro}

Using natural language in object detection to describe semantics bears the potential to significantly increase the size of the detector's label space and enable novel applications. While standard detectors operate on a fixed label space~\cite{ren_neurips15_fasterrcnn,COCO,Objects365}, natural language allows for a broad spectrum of object descriptions, ranging from generic terms like "vehicle" to specific expressions like "the red sports car parked on the left side"~\cite{schulter2023omnilabel,xie2023dcube,kamath2023tricd,liu2023gres,yu_eccv16_refcoco,mao_cvpr16_refcocog,li_22_elevater_odinw}. Several works advanced language-based object detection over the past few years with novel training strategies~\cite{gu_iclr_22,zhao2022exploiting,dou2022coarse_fiber,li2022grounded,lin2022learning_vldet,minderer_neurips23_owlv2,min2023neurocs} and model architectures~\cite{kuo2022f_iclr23_fvlm,minderer_eccv22_owlvit,subramanian2022acl_reclip,kamath_iccv_21}.

\begin{figure}[t]
  \centering
  \vspace{-2mm}
  \includegraphics[width=1.0\linewidth]{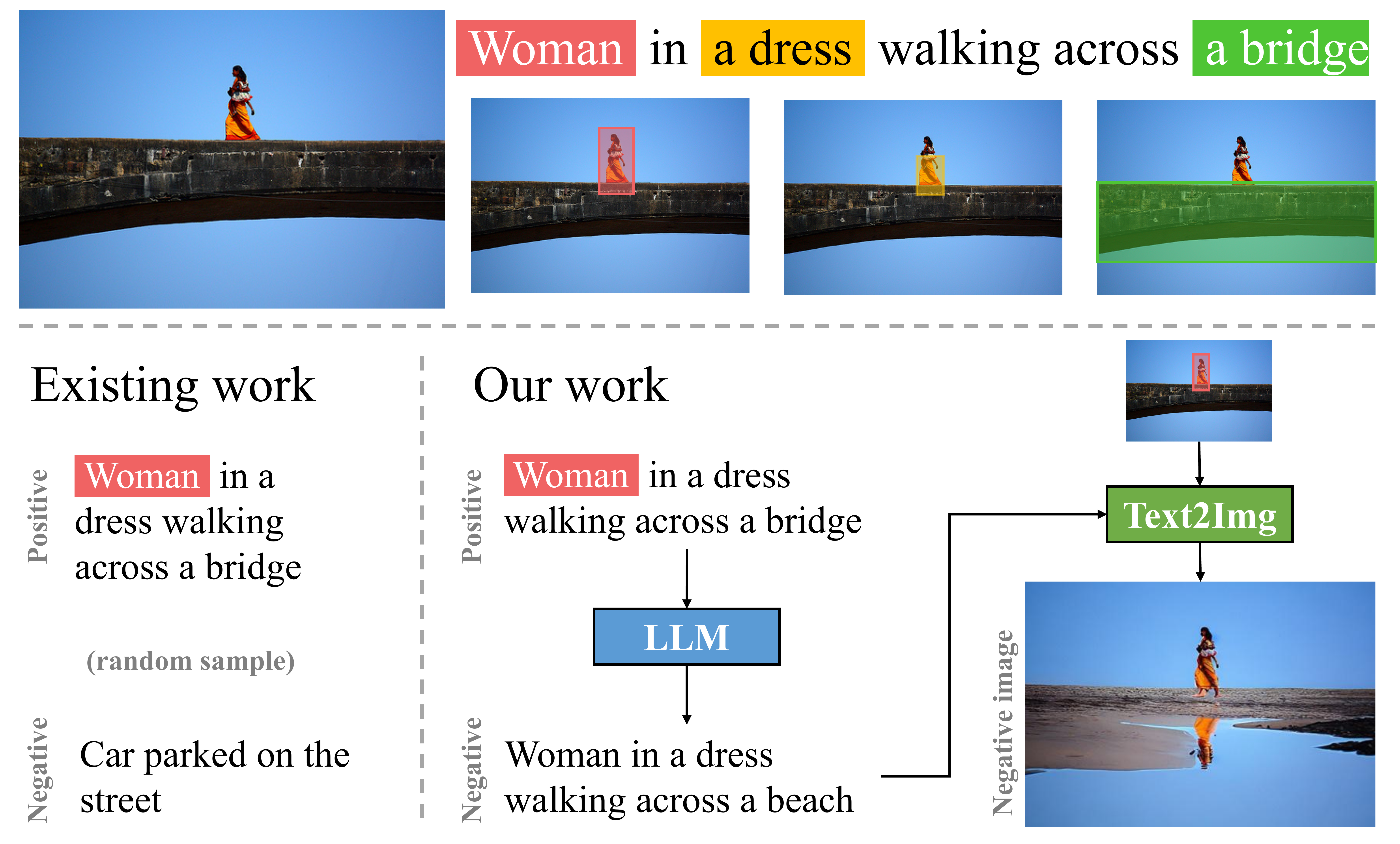}
  \caption{
  The key contribution of our work is to leverage large-language-models and text-to-image diffusion models to automatically generate negative object descriptions and images for training language-based object detectors. In contrast to prior work, our generated negatives are more relevant to the original data and provide a better training signal for detectors. 
  }
  \label{fig:teaser}
\end{figure}

Referring expression or visual grounding datasets~\cite{yu_eccv16_refcoco,mao_cvpr16_refcocog,wu2020phrasecut,plummer2015flickr30k,krishna2017visualgenome} provide the natural language object descriptions along with bounding box annotations needed for training. However, this data only describes what is present in the images, \emph{but misses to describe what is not}. Yet, the notion of negatives is crucial for training discriminative models like language-based detectors~\cite{robinson_iclr21,lin_iccv17_focalloss_retinanet,hinami_emnlp18_negativephraseaug}.

Detection datasets with a fixed label space provide negative classes implicitly or explicitly, with exhaustive~\cite{COCO,Objects365} or federated~\cite{OpenImages,gupta2019lvis} annotations, respectively. Any part of an image that does not overlap significantly with a bounding box of category $c$ is verified to \emph{not be of that category} (for exhaustive annotation). 
On the other hand, the space of negatives for a free-form text description of an object is extremely large. While some existing datasets provide negative samples in free-form text~\cite{shekhar_acl2017_foilcoco,thrush_cvpr2022_winoground}, they were not annotated with bounding boxes. Hence, existing language-based detectors often define the negatives for one object as the descriptions of all other objects in the same image or descriptions of other random samples~\cite{kamath_iccv_21,li2022grounded,dou2022coarse_fiber}.
However, such negatives may not be directly related to the original positive description and define a weaker training signal (see~\cref{fig:teaser}). By explicitly evaluating on human-curated negatives, a recent benchmark~\cite{schulter2023omnilabel} identified a bias of existing language-based detectors to perform clearly better on positive rather than negative descriptions. However, creating a dataset with high-quality human-curated negatives for large-scale training is labor-intense and costly.

In this work, we propose to explicitly and automatically \emph{generate} negative data in the form of free-form texts as well as images. Prior works~\cite{ma_cvpr23_crepe,doveh_2023_structured_vl_concepts,shekhar_acl2017_foilcoco,thrush_cvpr2022_winoground,hinami_emnlp18_negativephraseaug,yuksekgonul_iclr23_vlm_bagofwords} rely on rule-based approaches with knowledge graphs and focus only on the language domain or the classification task. In contrast, we leverage generative large-language-models (LLMs)~\cite{openai2022gpt-3.5,touvron2023llama} and text-to-image diffusion models~\cite{rombach2022high,li2023gligen} to automatically create relevant but contradicting object descriptions along with the corresponding images for language-based object detection, see \cref{fig:teaser}.

Given an object description of a dataset, we first use LLMs to generate a semantically contradicting description as the negative. Besides changing individual words (foils) based on explicit knowledge graphs or LLMs, like in prior work~\cite{hinami_emnlp18_negativephraseaug,li_neurips23_desco,doveh_2023_structured_vl_concepts}, we demonstrate improved detection performance with two alternative approaches. \emph{(Re-combination):} An LLM first identifies all objects in a sentence, and then creates a contradicting one by re-arranging, ignoring or adding objects. \emph{(In-context summaries):} We prompt an LLM to summarize the differences of a few (less than 100) positive-negative pairs collected from an existing image-level dataset~\cite{thrush_cvpr2022_winoground}. This summary is then used as context to generate more such examples.
Note that we do not need visual input for this step, allowing us to leverage powerful LLMs for semantic and textual reasoning.  Moreover, while prior work only focused on the text~\cite{hinami_emnlp18_negativephraseaug,shekhar_acl2017_foilcoco,ma_cvpr23_crepe,doveh_2023_structured_vl_concepts,yuksekgonul_iclr23_vlm_bagofwords}, we also leverage text-to-image diffusion models like GLIGEN~\cite{li2023gligen} to create images that match the generated negative descriptions of objects, which serves as additional training signal. While the direct output of such image-generation models is often noisy and even wrong (not matching the input description), we propose two filtering steps to reduce noise considerably (from 53\% to 16\% according to an empirical study). Having both negative object descriptions and the corresponding image, allows us to improve the discriminative loss for training language-based object detectors.

Our experiments demonstrate clear accuracy gains on two challenging benchmarks, +2.9AP on OmniLabel~\cite{schulter2023omnilabel} and +3.3AP on D$^3$~\cite{xie2023dcube}, when adding our automatically-generated negative data into the training of baseline models like GLIP or FIBER. Moreover, we provide an in-depth analysis of the generated data (text and images) and how they contribute to better language-based detection.

\noindent \textbf{Summary of contributions:}
(1) Automatic generation of semantically relevant but contradicting negative text and images with large-scale generative models. (2) Recipes to integrate such negative data into language-based detection models like FIBER~\cite{dou2022coarse_fiber} and GLIP~\cite{li2022grounded} (3) Clear improvements on language-based detection benchmarks~\cite{schulter2023omnilabel,xie2023dcube} including a thorough analysis of the generated data.

%% file: sec/2_related_work.tex
\section{Related work}
\label{sec:related_work}

\paragraph{Vision \& language localization tasks:} Open-vocabulary detection (OVD) requires a model to localize object category names without having seen explicit bounding box annotation for them~\cite{gu_iclr_22,zareian_cvpr_21,hinami_emnlp18_negativephraseaug,kuo2022f_iclr23_fvlm,wu2023cora,zhong2022regionclip}. In contrast, we focus on the more general language-based object detection task~\cite{schulter2023omnilabel,xie2023dcube}, which goes beyond simple category names. Referring expression comprehension (REC) aims at localizing the subject of a free-form text expression. However, REC benchmarks~\cite{yu_eccv16_refcoco,mao_cvpr16_refcocog,wu2020phrasecut} fall short in evaluating all aspects of the more general language-based detection task~\cite{schulter2023omnilabel,xie2023dcube}. In visual grounding (VG)~\cite{plummer2015flickr30k}, the task is to localize noun phrases of a caption in the image. Although being a task on its own, VG datasets have recently been used mostly as training data for OVD. Our work focuses on general language-based object detection, which subsumes and generalizes standard detection, OVD and REC~\cite{schulter2023omnilabel,xie2023dcube,kamath2023tricd,liu2023gres}.

\paragraph{Language-based object detectors:} Two critical abilities of language-based detection are accurate localization and tight text-image fusion. Works like \cite{Lu_neurips_19_ViLBERT,hu_iccv_21_UniT,Chen_ECCV_20_UNITER} use language-models like BERT~\cite{liu2019roberta,delvin19_bert} to align regions extracted from (pre-trained) detectors with captions. The outstanding zero-shot classification accuracy of large-scale pre-trained models like CLIP~\cite{radford_icml_2021_CLIP} or \cite{jia_icml_21,khan_eccv22_simla,ALBEF} then sparked interest in extensions for localization, with different approaches like distillation~\cite{gu_iclr_22}, fine-tuning~\cite{kuo2022f_iclr23_fvlm,minderer_eccv22_owlvit}, pseudo-labeling~\cite{zhao2022exploiting,zhao2023sasdet,minderer_neurips23_owlv2}, or combinations thereof~\cite{li2022grounded,dou2022coarse_fiber}.
We use such models as test bed, but explore the underlying training data with respect to negative samples.

\paragraph{Negative samples for object detection:} The notion of negatives is crucial for training discriminative models~\cite{robinson_iclr21,lin_iccv17_focalloss_retinanet}. Also for object detection, hard negative mining~\cite{shrivastava_cvpr16_ohem} has proven beneficial for model training. However, these prior works aim to find hard negative training examples rather than negatives in the label space, because the label space is fixed in standard detection. For language-based datasets, the space of potential negatives is extremely large because object descriptions are free-form text.  Prior works~\cite{doveh_2023_structured_vl_concepts,ma_cvpr23_crepe,yuksekgonul_iclr23_vlm_bagofwords,thrush_cvpr2022_winoground,shekhar_acl2017_foilcoco} investigate negative texts for general vision \& language models with different strategies, including changing individual words (foil) with rules based on knowledge graphs~\cite{wordnet} or with LLMs. 
SugarCrepe~\cite{hsieh2024sugarcrepe} shares a similar idea as us to get negative texts with in-context learning but for image-text level pretraining. 
For language-based detection, \cite{hinami_emnlp18_negativephraseaug} explores such rule-based foils, while \cite{li_neurips23_desco} uses LLMs with specific templates to replace object names with alternative descriptions.
In contrast, our work (1) focuses on the localization task, (2) explores more comprehensive strategies to generate negatives with LLMs, and (3) proposes to also generate corresponding negative images with text-to-image diffusion models.

%% file: sec/3_method.tex
\section{Method}
\label{sec:method}

\begin{figure}[t]\centering
  \includegraphics[width=1.0\linewidth]{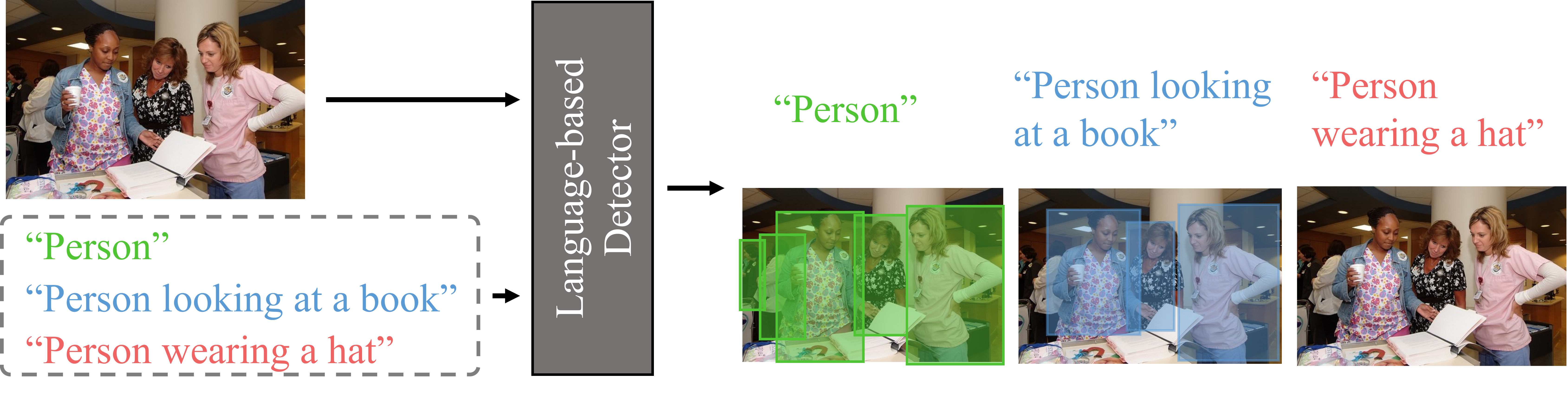}
  \caption{
  In language-based object detection, a detector receives as input an image and a (variable-length) list of free-form text descriptions of objects. For each description, the model predicts bounding boxes for objects that match the description.
  }
  \label{fig:overview_task}
\end{figure}

\subsection{Language-based object detection}
\label{sec:method_language_based_detection}

\paragraph{Task definition:} Given an image and a list of object descriptions, the task is to output bounding boxes along with confidence scores for each description, as shown in \cref{fig:overview_task}. Note the multi-label setting where one object instance can be referred to by multiple descriptions, like ``person'' and ``person looking at book''. Also note that an object description might refer to zero objects in the image and the desired output is an empty set of boxes.

\paragraph{Training data:} Many language-based detection models~\cite{li2022grounded,dou2022coarse_fiber} use a combination of object detection~\cite{COCO,Objects365} and visual grounding datasets~\cite{plummer2015flickr30k,krishna2017visualgenome} for training their models. Both types of datasets provide images $I$ and bounding boxes $b_l$ to localize individual objects. Object detection data assigns each bounding box $b_l$ a unique category $c$ out of fixed label space $\mathcal{C}$. The exhaustive labeling of the fixed label space in detection datasets implies the space of negatives. An object of category $c$ is not any of the categories $\mathcal{C} \setminus c$. On the other hand, grounding data provides an image caption $t$ in free-form text, where subsets of words $m_l$ (defined as indices of starting and ending characters in $t$) are linked to bounding box $b_l$.
For grounding data, the space of negatives is extremely large because one can find as many textual descriptions that do not match $t$ as desired due to the compositionality of free-form text. Many language-based detectors~\cite{li2022grounded,dou2022coarse_fiber} only use the words in $t$ that are not referred to by $m_l$ as negatives for the bounding box $b_l$. We argue that this choice is sub-optimal because these words may refer to entirely different objects and are easy to discriminate. In the following section, we explain how we can automatically generate negative samples that semantically related to the original text $t$ and hence provide a better training signal.

\subsection{Generating negative samples}
\label{sec:method_generate_negative_samples}
Our goal is to automatically and explicitly generate negative samples based on the original text descriptions $t$ to improve the training signal for language-based detectors. A key observation of our work is to leverage the vast knowledge encoded into large-language models (LLMs)~\cite{openai2022gpt-3.5,touvron2023llama} and text-to-image diffusion models (T2I)~\cite{rombach2022high,li2023gligen}. Besides proposing novel ways to instruct LLMs for generating negative text descriptions (\cref{sec:method_gen_neg_text}), we also propose to generate negative images (\cref{sec:method_gen_neg_images}).

\subsubsection{Generating negative descriptions with LLMs}
\label{sec:method_gen_neg_text}
Given an object description $t$ that matches the visual content inside a bounding box $b_l$, we define a ``negative'' description $t'$ as any text that is \emph{semantically different} to the original text. Furthermore, our intuition is that good negative descriptions are still semantically related to the original description, but not the same. An example is: ``Person in red shirt'' as the original description and ``Person in blue shirt'' as a contradicting negative one.

Prior work~\cite{shekhar_acl2017_foilcoco,doveh_2023_structured_vl_concepts,hinami_emnlp18_negativephraseaug} explored rule-based approaches to generate negative text. However, such rules are typically limited to simple knowledge graphs and are limited to replacing only individual words, often just nouns, or swapping words. 
In contrast, we explore more powerful LLMs to automatically generate relevant negatives.  
To make the negative text generation efficient and economic, in all cases, we first leverage a strong instruction-tuned LLM~\cite{openai2022gpt-3.5} to generate 50k positive-negative pairs, and then finetune a LLaMA-7B~\cite{touvron2023llama} model with those pairs to then generate negative captions on large grounding datasets. 
In the following, we describe three ways to instruct an LLM for generating positive-negative pairs of object descriptions:

\paragraph{LLM-based foils:} 
We first prompt an instruction-tuned LLM~\cite{openai2022gpt-3.5} to find concepts (i.e., objects, attributes and relationships) in object descriptions. Compared to rule-based parsers~\cite{wu_cvpr19_parser}, LLMs can provide richer information.  For example, for the caption ``A transportation vehicle is carrying a crowd of people who are sitting and standing.'', the parser ignores ``sitting'' and ``standing'', while LLMs regard them as attributes. Then, we pick one concept from the first step sequentially and prompt LLMs again to generate a negative caption by changing the concept. For both steps, the prompts are manually curated with the task definition and step-by-step instructions for the generation. Please find the exact prompt for the LLM in the supplement. 

\paragraph{Re-combination:} Next, we give the LLM more freedom in generating negative descriptions. 
We first prompt the LLM to identify all objects in the original caption, and then to re-combine them to create a new sentence different from the original one. We allow the LLM to ignore, change or add new objects. For example, given the caption ``A boy is playing with his dog'' and two objects ``boy'' and ``dog'', the LLM can output ``The girl and her dog are playing fetch in the park''. 
Detailed prompts for both identifying objects and re-combination are in the supplement.

\paragraph{In-context summary:} 
Third, we enable LLMs to learn how to generate negative descriptions by providing human-annotated positive-negative pairs as in-context samples. 
We randomly sample 80 pairs of positive and negative texts from the Winoground dataset~\cite{thrush_cvpr2022_winoground} and prompt the instruction-tuned LLM~\cite{openai2022gpt-3.5} to summarize the difference of those pairs in plain text.  
Then, instead of manually creating prompts to generate positive-negative pairs, we leverage the summary together with three randomly sampled Winoground pairs as prompts to the LLM, and generate several positive-negative pairs to finetune LLaMA.
After finetuning, the LLaMA model is used to generate negative texts for given descriptions.
This pipeline does not require hand-crafted prompts to LLMs as the explanation of the concept of negatives and how to create them. 
The supplement contains full prompts for generating a summary, and generating positive-negative pairs for finetuning.

\subsubsection{Generating negative images with T2I models}
\label{sec:method_gen_neg_images}
Given an original image $I$, a bounding box $b$ and a corresponding object description $t$, we define a negative image $I'$ as any image that has a different semantic content inside $b$. The rest of the image can be equivalent to $I$. To obtain such imagery, we start with visual grounding data that provide bounding boxes, positive captions with text phrases, and alignment between them. We propose a two-step process: First, we turn the positive caption into a negative one. Second, we use conditioned image generation tools to alter the visual content inside the bounding box $b$.

\begin{figure}\centering
  \includegraphics[width=1.0\linewidth]{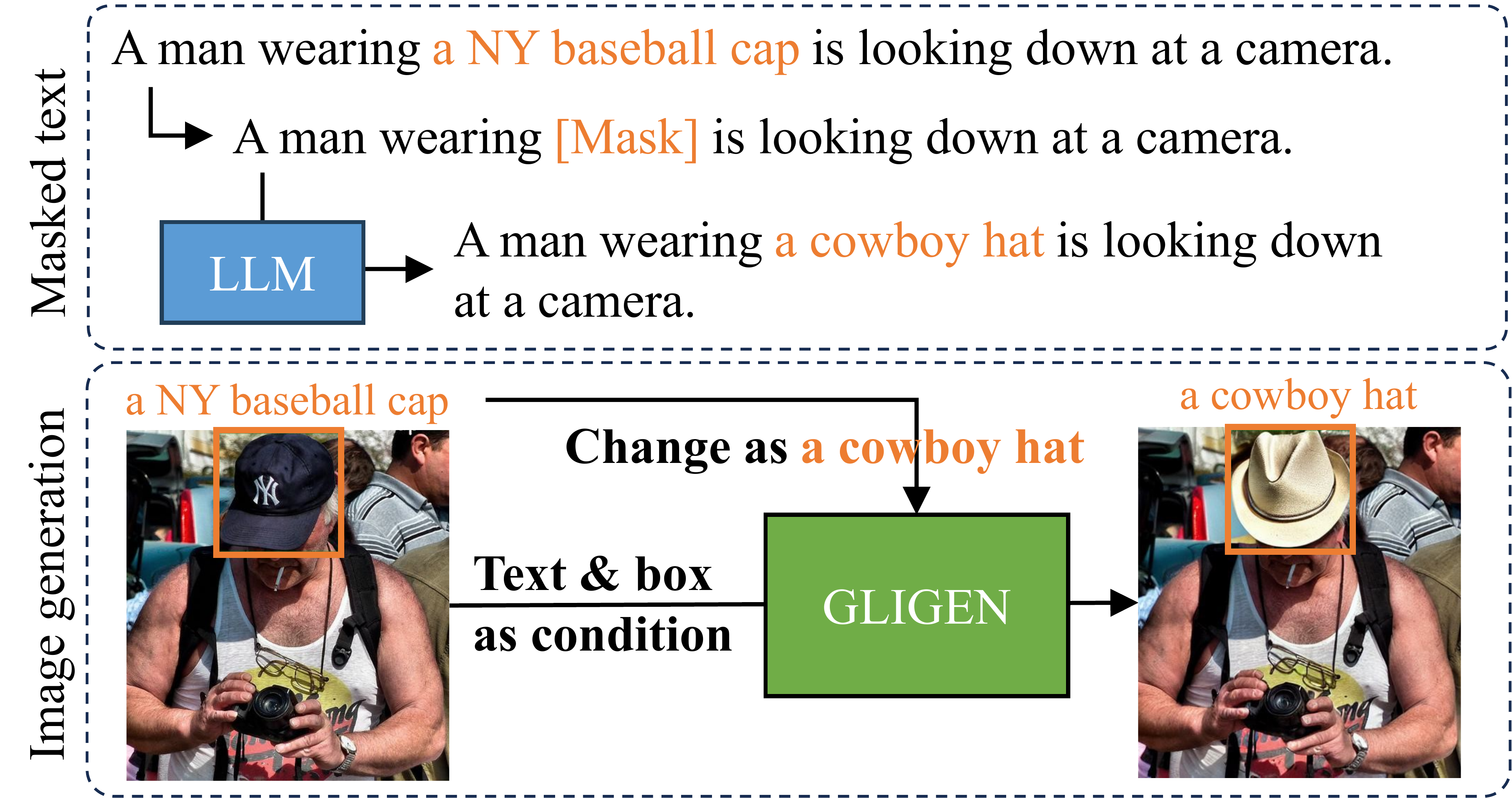}
  \caption{Overview of using LLMs~\cite{openai2022gpt-3.5,touvron2023llama} and text-to-image diffusion models~\cite{rombach2022high,li2023gligen} to generate negative images.}
  \label{fig:gen_neg_img}
\end{figure}

\paragraph{Negative text for negative images:} Although we have already described an approach to generate negative descriptions in \cref{sec:method_gen_neg_text}, doing so to generate a negative image requires a different approach. In this case, the generated negative text needs to preserve the alignment $m_l$ to the ground truth bounding box $b_l$ in order to instruct the generative image model GLIGEN~\cite{li2023gligen}. Hence, we first select a bounding box $b_l$ and mask out the corresponding words (known via $m_l$) in the text $t$. For example, ``A boy is playing with his dog'' turns into ``A boy is playing with [Mask]'' if the selected bounding box refers to ``his dog''.
Again, we leverage LLMs~\cite{openai2022gpt-3.5} to fill in text for ``[Mask]'' to generate a negative text without reusing the original text. Please refer to Fig.~\ref{fig:gen_neg_img} for illustrations.

We finetune a LLaMA-7B for the mask filling task with triplets of positive texts, masked texts, and negative texts.
To reduce manual efforts, we follow the approach of in-context summary to get triplet samples.
We apply this process twice: We start with only 5 manually created triplets to build a summary and generate 100 samples from the LLM~\cite{openai2022gpt-3.5} with human checks. 
We then repeat the process to generate 50k examples without human checks from a summary of the 100 generated examples. 
This increases diversity in the generated data.

\paragraph{Conditional image generation:} Given an image $I$, a bounding box $b$ and the altered text $t'$, we generate a negative image $I'$ that is equal to $I$ except inside $b$, where the visual content is altered to match the text $t'$. To do so, we use the inpainting and conditioning abilities of GLIGEN~\cite{li2023gligen}, a T2I model~\cite{rombach2022high}. Refer to \cite{li2023gligen} and our supplemental material for more details, and to \cref{fig:gen_neg_img} for an illustration of the process.

\paragraph{Mitigating noise in image generation:} We found that the generated images are often noisy for any of the following reasons: 
(1) The altered text refers to a big bounding box that covers other smaller boxes. Large portions of the image are then generated and often do not match the concepts those smaller boxes originally covered. 
(2) The generated negative text does not match the bounding box that is either too small, too large or in a inappropriate position.
(3) The T2I model fails to understand the negative text and generates wrong content. We propose two steps to filter such noisy images. First, we simply ignore ground truth boxes $b_l$ for image generation if the box covers more than 75\% of any other boxes in the image. Second, we adopt CLIP~\cite{radford_icml_2021_CLIP} to verify the semantic similarity of the generated image regions and the corresponding text. Specifically, we compute the similarity with CLIP between the generated image region (visual input) and the original and generated negative texts (text input). We filter out generated images that have a similarity score to the generated negative text lower than a user-defined threshold. 
Details on the filtering steps are given in the supplemental.

\begin{figure}\centering
  \includegraphics[width=1.0\linewidth]{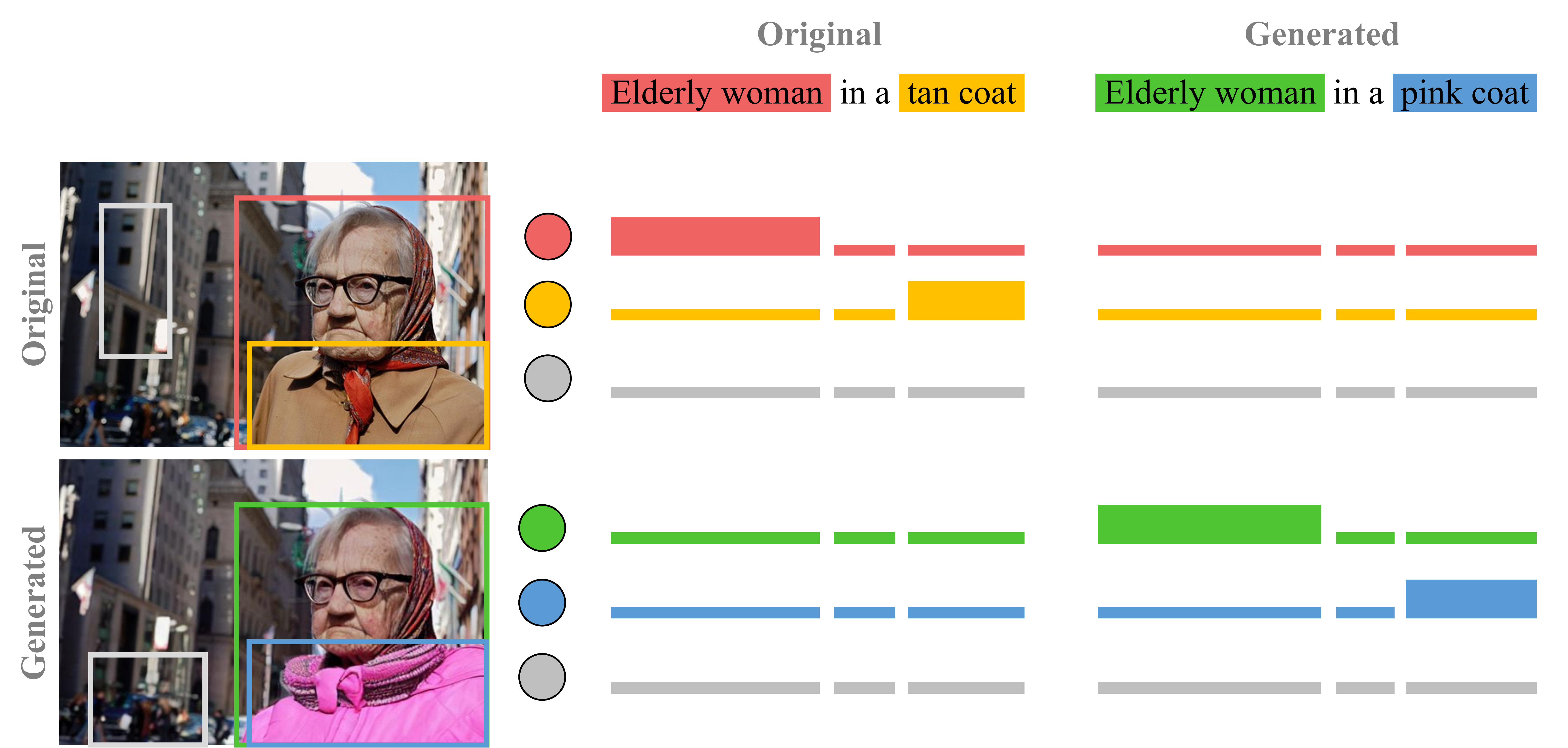}
  \caption{
  Illustration of the grounding loss used for training.
  Predictions that are matched with ground truth receive a positive signal from the associated text (tall rectangles). All other words receive a negative signal (short rectangles). The top left quarter shows the original loss used in~\cite{dou2022coarse_fiber,li2022grounded}. The other three quarters are related to our proposed \emph{generated} negative data and provide additional training signals.
  }
 \label{fig:model_loss}
\end{figure}

\subsection{Learning from negative samples}
\label{sec:method_learning_from_negative_samples}

\paragraph{Detector design and training objective:} The generated data does not prescribe any specific architecture for the detector. A common choice, which we also use for our experiments, is \cite{li2022grounded,dou2022coarse_fiber}. The inputs are image $I$ and text $t$, and the output is a set of bounding boxes $\hat{b}_i$ with corresponding logits $\hat{p}_i \in \mathbb{R}^T$. Here, $T$ is the number of tokens required to represent text $t$. The ground truth can be represented by a binary assignment matrix $\mathbf{A} \in \mathbb{B}^{L \times T}$. Rows refer to ground truth boxes $l$ and columns to tokens in $t$. Each element indicates if a token corresponds to a box $l$, which is given by the ground truth indices $m_l$. To define a loss, bipartite graph matching associates predictions with ground truth. For matched predictions, the target vector $g_i \in \mathbb{B}^T$ is the corresponding row from $\mathbf{A}$, while it is an all-zero vector for unmatched targets. 
The loss is then computed as $\mathcal{L} = \sum_i \ell_{\textrm{FL}} \left( \hat{p}_i, g_i \right)$ where $\textrm{FL}$ refers to a focal loss.
\cref{fig:model_loss} illustrates the loss.

\paragraph{Integrating negative text:} When sampling an image $I$, along with text $t$, boxes $b_l$ and indices $m_l$, we also randomly sample $K > 1$ negative descriptions from $\{t'_j\}$ that defines the pool of negatives generated for text $t$. We randomly shuffle the order of all texts to avoid any biases on the location of the one positive description, and then concatenate them into one text string.

\paragraph{Integrating negative images:} We explore two options: (1) Simply add the generated images $I'$ along with their generated (but semantically matching) captions $t'$ as additional visual grounding data. The original caption $t$, which was the starting point to generate the negative image $I'$, is now used as the negative caption. In this way, both the original image $I$ and the generated one $I'$ have positive and negative descriptions. This option is illustrated in \cref{fig:model_loss}.  (2) To better leverage the relation between the original and generated data, the second option is to pack them into a single training sample. We simply concatenate the images $I$ and $I'$, as well as the texts $t$ and $t'$. The ground truth information $m_l$ is updated accordingly. See supplement for details.

%% file: sec/4_experiment.tex
\section{Experiments}
\label{sec:experiment}

\subsection{Experimental design}
\label{sec:exp_experimental_design}

\paragraph{Training procedure:} We choose two recent methods, GLIP-T~\cite{li2022grounded} and FIBER-B~\cite{dou2022coarse_fiber}, to demonstrate the effect of our automatically generated negatives. We use the official code and publicly available checkpoints as a starting point. The Flickr30k dataset~\cite{plummer2015flickr30k} serves as our grounding dataset to generate the negative data. 
We then fine-tune GLIP-T and FIBER-B with both positive and negative data, along with the Objects365 detection dataset~\cite{Objects365} for 1 epoch. 
Note that both Objects365 and Flickr30k are part of the original training set. We do not introduce any extra data except our generated negatives.
Most hyper-parameters are equal to the original settings of GLIP and FIBER. Any exceptions are described in the supplement.

\paragraph{Evaluation benchmarks:} We choose two recently proposed benchmarks, OmniLabel~\cite{schulter2023omnilabel} and D$^3$~\cite{xie2023dcube}, as our test beds. These benchmarks evaluate more aspects of language-based detection than existing referring expressions~\cite{yu_eccv16_refcoco,mao_cvpr16_refcocog,wu2020phrasecut} or open-vocabulary detection~\cite{li_22_elevater_odinw,gupta2019lvis} benchmarks. Specifically, both benchmarks contain complex object descriptions that go beyond simple category names from open-vocabulary detection benchmarks. Moreover, the descriptions can refer to zero, one or multiple instances in the image, in contrast to standard referring expression benchmarks. These properties enable a more stringent evaluation metric as in object detection, which is based on average precision (AP) in both OmniLabel~\cite{schulter2023omnilabel} and D$^3$~\cite{xie2023dcube}.
Both benchmarks provide more fine-grained metrics. OmniLabel evaluates separately for categories, descriptions, and descriptions referring to at least one object, with APc, APd and APd-P, respectively. 
D$^3$ differentiates descriptions on absence (``Abs'') and presence (``Pres'') that indicate whether or not they contain any form of negation (e.g., ``without''), as well as on text lengths. 
Finally, we create a specific split for OmniLabel, ``OmniLabel-Negative'', to evaluate the model only on images that contain at least one negative description (\ie, not referring to any object).

\input{tbls/tbl_main_results_ol}
\input{tbls/tbl_main_results_d3}

\subsection{Benchmark comparisons}

\cref{tab:main_results_ol,tab:main_results_d3} evaluate the impact of our generated negative training data on the OmniLabel~\cite{schulter2023omnilabel} and D$^3$~\cite{xie2023dcube} benchmarks. In both tables, the first set of rows are baselines provided by the benchmarks. The following rows show the main comparisons for GLIP-T~\cite{li2022grounded} and FIBER-B~\cite{dou2022coarse_fiber} with and without adding our generated negative training data.
First, we can see that adding negative data improves results across all metrics for both models and both benchmarks. On OmniLabel, we can see a +2.9\% and +2.4\% increase in AP for GLIP-T and FIBER-B, respectively. Similarly, we observe a +2.3\% and +3.3\% increase in the main metric of D$^3$ (AP on full descriptions) for GLIP-T and FIBER-B.

\begin{table*}
\parbox[t]{.60\textwidth}{\null
\centering
 \resizebox{.60\textwidth}{!}{ 
 \begin{tabular}{@{} l cccc cccc ccc @{}}
 \toprule
  & \multicolumn{4}{c}{\textbf{Whole OmniLabel}} & \multicolumn{4}{c}{\textbf{OmniLabel-Negative}} & \multicolumn{3}{c}{\textbf{D$^3$}} 
 \\ 
 \cmidrule(r){2-5} \cmidrule(lr){6-9} \cmidrule(lr){10-12}  

  & \textbf{AP} & \textbf{APc} & \textbf{APd} & \textbf{APd-P}
  & \textbf{AP} & \textbf{APc} & \textbf{APd} & \textbf{APd-P}
  & \textbf{Full} & \textbf{Pres} & \textbf{Abs}
  \\ 
 \midrule

 Original FIBER-B
     & 25.7 & 30.3 & 22.3 & 34.8
     & 18.7 & 31.2 & 13.3 & 36.3
     & 22.7 & 21.5 & 26.0
      \\
 + Rule-based foils 
     & 26.4 & {\bf 31.7} & 22.6 & 34.9
     & 19.2 & 32.6 & 13.6 & 36.4
     & 24.1 & 23.2 & 26.9
     \\
 + LLM-based foils
     & 26.5 & 30.7 & 23.3 & {\bf 35.9}
     & 20.8 & 32.1 & 15.4 & \textbf{38.0} 
     & 24.6 & 24.0 & 26.5
     \\
 + Re-combination
     & {\bf 26.9} & \underline{30.8} & \textbf{23.9} & {\bf 35.9}
     & \textbf{21.1} & \textbf{32.3} & \textbf{15.6} & \underline{37.6} 
     & \underline{25.3} & \underline{24.6} & \underline{27.3}
     \\
 + In-context summary
     & \underline{26.6} & \underline{30.8} & \underline{23.4} & 34.2
     & \textbf{21.1} & \underline{32.2} & \underline{15.7} & 36.4 
     & \textbf{25.7} & \textbf{25.2} & \textbf{27.5}
     \\ 

 \bottomrule
 \end{tabular}
 }
  \caption{Performance of FIBER-B trained with negative texts from four negative generation methods.}
 \label{tab:ablation_results_negtxt_ol}
}
\hfill
\parbox[t]{.375\textwidth}{\null
\centering
 \resizebox{.375\textwidth}{!}{ 
 \begin{tabular}{@{} l ccc ccc @{}}
 \toprule
& \multicolumn{3}{c}{\textbf{OmniLabel}} & \multirow{2}{*}{\textbf{D$^3$}}
 \\ 
 \cmidrule(r){2-4} 

   & \textbf{APc} & \textbf{APd} & \textbf{APd-P}
   & 
  \\ 
 \midrule

 Original FIBER
     & 30.3 & 22.3 & 34.8
     & 22.7 
     \\
\hline
 FIBER w/ neg. texts
     & 30.7 & 23.9 & 35.5
     & \underline{25.9}
     \\
 + W/ neg. img. directly
    & 30.1 & 22.4 & 33.7
     & 23.0
     \\
 + Box filter
     & 31.0 & 23.8 & 35.4
     & 23.6
     \\
 + Box\&CLIP filters
     & 31.1 & 24.2 & \underline{35.9}
     & 24.1
     \\ 
 + Above + concat. img.
     & \underline{31.7} & \underline{24.8} & \underline{35.9}
     & 24.8
     \\ 
 + Above + weight ensemble
     & \textbf{32.1} & \textbf{25.2} & \textbf{36.5}
     & \textbf{26.0}
     \\ 
 \bottomrule
 \end{tabular}
 }
\caption{FIBER trained with negative images.}
\label{tab:ablation_results_negimg_ol}
}
\end{table*}

\subsection{Analysis on negative texts}

{\bf Effectiveness of different negative texts:} 
We finetune  FIBER-B without and with different kinds of negative texts mentioned in Sect.~\ref{sec:method_gen_neg_text}, i.e., Rule-based foils, LLM-based foils, Re-combination with LLMs, In-context summary with LLMs, and present results in Table~\ref{tab:ablation_results_negtxt_ol}.
We find all kinds of negatives improve the original FIBER-B on both OmniLabel and D$^3$ benchmarks. 
Negative texts from LLMs generally achieve better results compared to LLM-based foils, which indicates that LLMs are powerful tools for negative text generation.
Moreover, both recombination and in-context summary with LLMs outperform LLM-based foils in all metrics except APd-P. Note that APd-P refers to evaluations without negative label spaces, which is a task weaker than language-based detection.
Based on such results, we argue that although word foils provide promising results in traditional studies~\cite{hinami_emnlp18_negativephraseaug,shekhar_acl2017_foilcoco}, it is sub-optimal to LLMs.
We need to explore varied ways to unlock the ability of LLMs.
We believe that our two solutions, i.e., Re-combination and In-context summary, provide a good starting point for future studies.
Besides using only one kind of negative texts, we also explore the combinations of different kinds of negative texts in the supplement.

\begin{figure}[tb]
  \centering
  \includegraphics[width=1.0\linewidth]{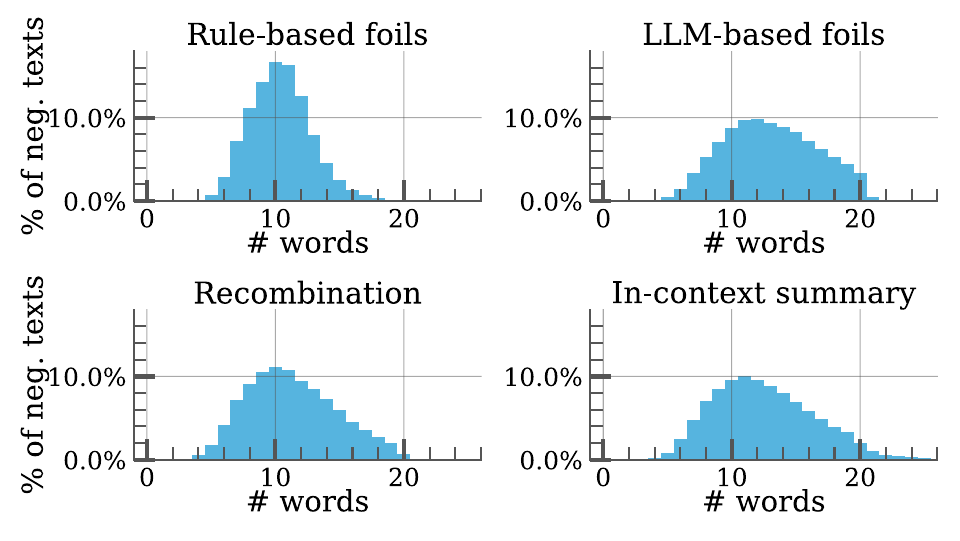}
  \vspace{-5mm}
  \caption{Percentage of negative texts with the numbers of words. Four negative generation methods are compared.}
  \label{fig:hist_sent_len}
\end{figure}

\begin{figure}[tb]
  \centering
  \includegraphics[width=1.0\linewidth]{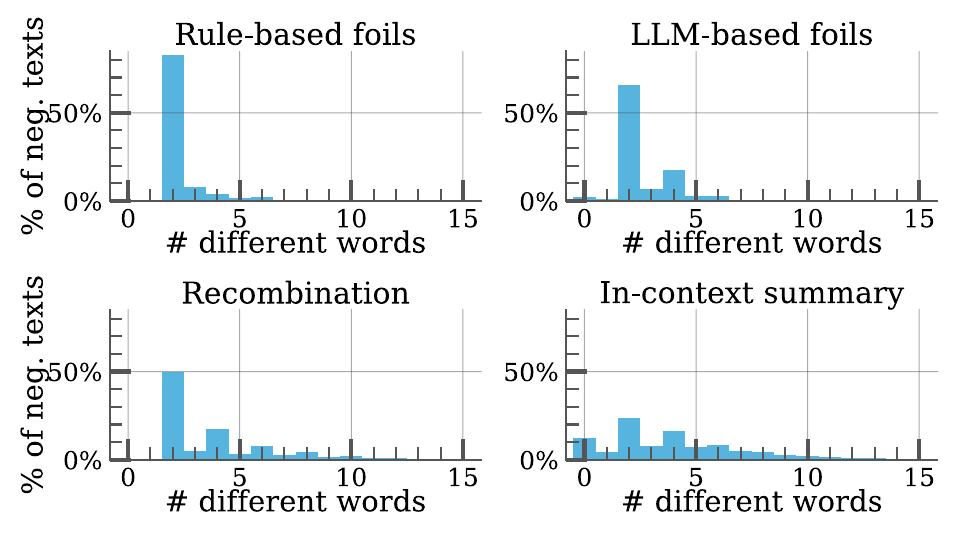}
  \vspace{-5mm}
  \caption{Percentage of negative texts with the numbers of words that are different from the original caption. Four negative generation methods are compared.}
  \label{fig:hist_num_of_diff_words}
\end{figure}

\begin{figure}[tb]
      \centering
  \includegraphics[width=.75\linewidth]{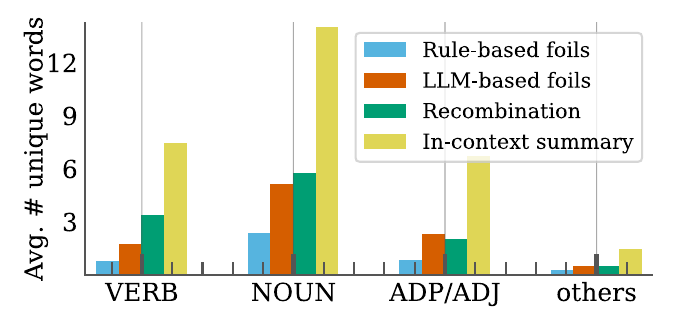}
  \vspace{-4mm}
  \caption{Average numbers of extra unique words per thousand generated negative texts, which are not included in the original dataset. We group words by their part-of-speech.}
  \label{fig:stats_unique_words_diff_negTxt}
\end{figure}

\paragraph{Diversity of rule-based and LLM-based negatives:}
In this part, we investigate the diversity of different negative texts.
First, we count number of words for each negative text and provide the distribution for negatives of different sources in Fig.~\ref{fig:hist_sent_len}.
As shown, all four distributions have a peak around 10 words, but the one of rule-based foils is higher than others. That means rule-based foils provide more negative texts with similar lengths.

Second, we count the number of different words between the original positive caption and the negative caption, and present the distributions in Fig.~\ref{fig:hist_num_of_diff_words}.
We find that LLM-based methods usually changes more words than rule-based foils, which increases the diversity. 
Moreover, in-context summary has a more flat distribution compared to others. Probably, in-context summary learns how to generate negatives automatically from data and has less restrictions.
Besides, in-context summary has more cases with no word changed where negative texts are generated by just shuffling words or concepts in the original text. 
Such shuffling is a common pattern of Winoground~\cite{thrush_cvpr2022_winoground}, and our in-context summary can learn such data specific patterns.

Third, we count how many extra words that does not exist in the original Flickr30k dataset are introduced in different negative generation methods.
Fig.~\ref{fig:stats_unique_words_diff_negTxt} shows the average number of extra words per 1000 negative texts.
We group words into four part-of-speech categories, i.e., VERB, NOUN, ADP/ADJ, and others.
As shown, LLMs introduce more extra words on average than rule-based foils probably because rule-based foils are limited in a predefined set of words. However, LLMs are open to any concepts and have great potentials of generating diverse texts.
In-context summary introduces the most extra words for all categories, which is likely a benefit of learning negative generation from data.
The above statistics indicate a clear view that LLMs generate more diverse data than rule-based foils.

\begin{figure}[tb]
  \centering
  \includegraphics[width=.75\linewidth]{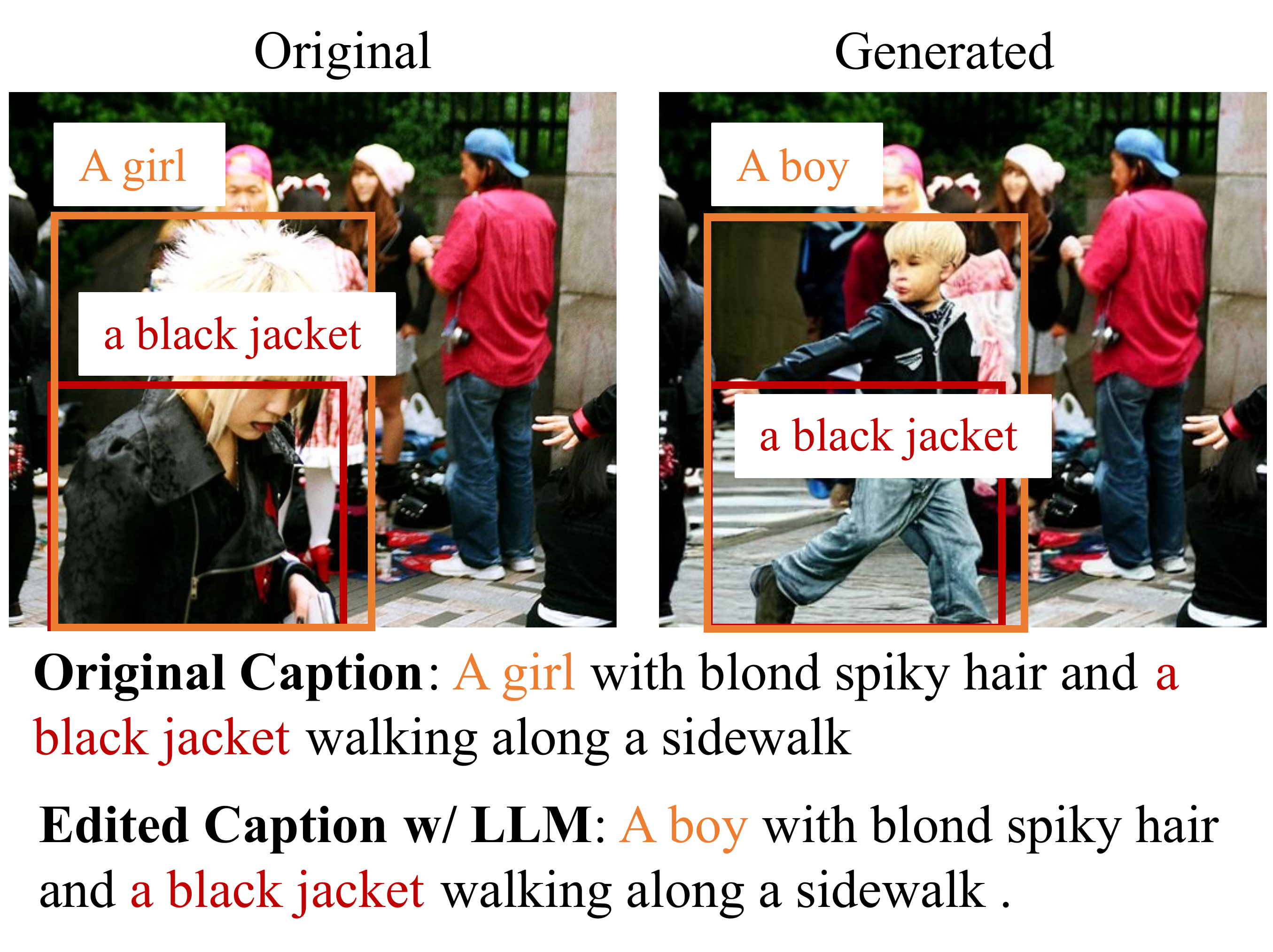}
  \caption{Noisy generated images. The orange box contains the red box, and editing the orange changes the red unexpectedly.}
  \label{fig:negImg_fail_large_box}
\end{figure}

\begin{figure*}[tb]
  \centering
  \includegraphics[width=.9\linewidth]{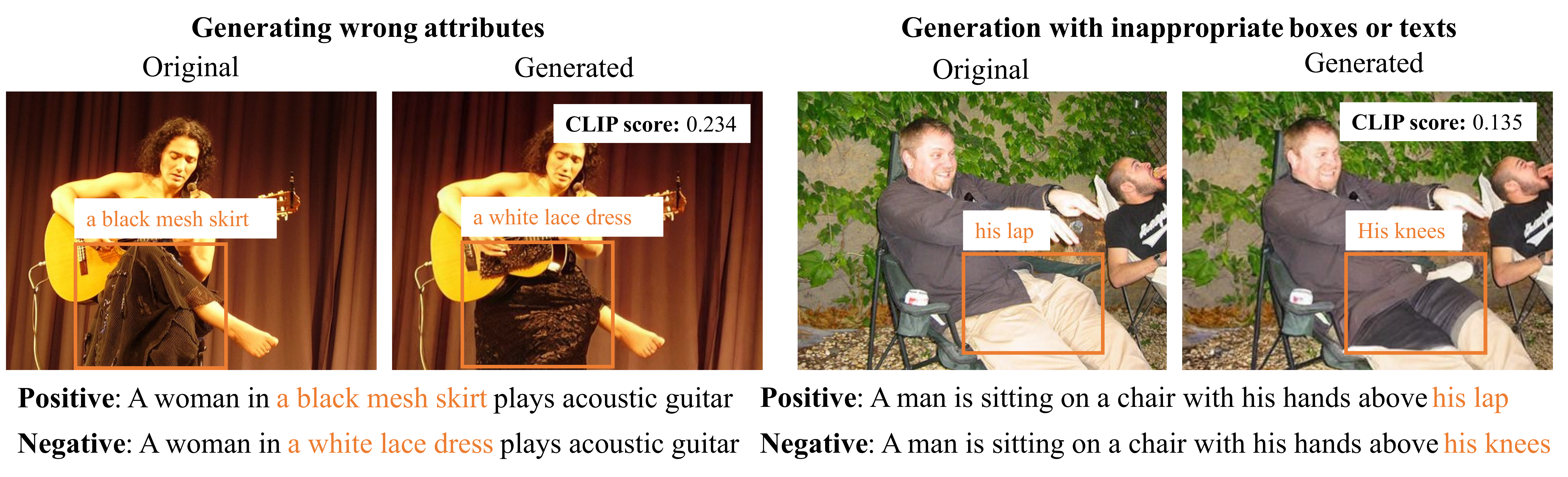}
  \caption{
{\bf Left:} Noisy negative images due to wrong attributes or objects generated by text-to-image models. 
{\bf Right:} Noisy negative images caused by inappropriate bounding boxes or negative texts from LLMs.
CLIP scores of generated images refer to the similarity between the box and the negative text compared to the positive text. Thresholding on CLIP scores remove those noisy images.
}
  \label{fig:negImg_fail_CLIP}
\end{figure*}

\begin{figure}[tb]
  \centering
  \includegraphics[width=.75\linewidth]{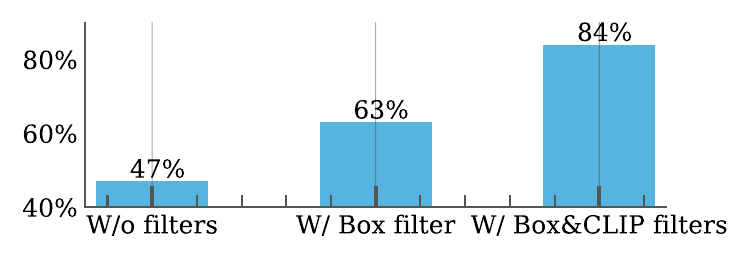}
  \vspace{-3mm}
  \caption{Percentage of good generated negative images.}
  \label{fig:negImg_subject_study}
\end{figure}

\subsection{Analysis on negative images}

{\bf Noise in generated images:}
As mentioned in the last paragraph of Sect.~\ref{sec:method_gen_neg_images}, the raw generated images are noisy in several ways.
First, the editing of a large box will override the context of smaller boxes that are covered by the large box. 
As shown in Fig.~\ref{fig:negImg_fail_large_box}, GLIGEN did follow the instruction to generate a boy in the orange box, but the black jacket in the red box is missing.
As a remedy, we apply our first de-noise step  ``Box Filter''. That is, we ignore boxes that contain any other boxes when generating negative images 
Second, GLIGEN may generate contents with wrong attributes or objects, as shown in Fig.~\ref{fig:negImg_fail_CLIP} (Left).
Moreover, our generation pipeline includes some cases where the edited text and the bounding box does not match. 
As shown in Fig.~\ref{fig:negImg_fail_CLIP} (Right), the box for ``his lap'' cannot be modified as ``his knees''. 
Thus, GLIGEN generates wrong contents.
As described in Sect.~\ref{sec:method_gen_neg_images}, we adopt a pretrained CLIP model to judge if generated contents are correct, which mitigates the noise to some extent.
As shown in Fig.~\ref{fig:negImg_fail_CLIP}, both negative images get low CLIP scores and can be filtered out with a threshold. We call such thresholding ``CLIP Filter''.

\paragraph{Subject studies on Box and CLIP filters:}
We employ human experts to check the amount of noisy generated images. 
First, for negative images w/o filter, w/ Box filter, and w/ Box\&CLIP filters, we separately and randomly select 100 samples.
Then, we ask two experts to check if a negative image is not noisy by comparing it with its caption and the original positive image.
We regard an image as not noisy when both experts agree.
As shown in Fig.~\ref{fig:negImg_subject_study}, both filters reduce the noise. The Box filter improves from 47\% to 63\%, and the CLIP filter improves to 84\%.

\paragraph{Effectiveness of generated negative images:}
To show the effectiveness of generated images themselves, we take captions of generated images as additional negative texts to in-context summary, and finetune a FIBER model as baseline.
Then, we compare the baseline with variants of adding generated negative images in Table~\ref{tab:ablation_results_negimg_ol}.
As shown, the performance drops if we directly take raw negative images as new visual grounding data without any filters (i.e., W/ neg. img. directly). 
Probably, there are too much noise in raw negative images as shown in Fig.~\ref{fig:negImg_subject_study}.
When applying both Box and CLIP filters on negative images, we can achieve slight improvement on OmniLabel compared to using negative texts only.

\paragraph{Concatenating images during training:}
Following the idea of concatenating the positive and negative captions as the text input, we concatenate the positive and negative images as one input image during training. See supplement for an example.
In this way, models are forced to tell the difference between the positive and negative images within one training iteration, which helps detectors to learn better about the negative.
As shown in Table~\ref{tab:ablation_results_negimg_ol}, such a simple technique improves upon ``+ Box\&CLIP filters'' both on OmniLabel and D$^3$.
Furthermore, we ensemble the weights of two FIBER models, one finetuned with negative texts only, and the other finetuned with both negative texts and images. 
Finally, compared to using negative texts only, we gain 1.3 APd on OmniLabel and no performance drop on D$^3$.

\begin{figure}[tb]
  \centering
  \includegraphics[width=1.0\linewidth]{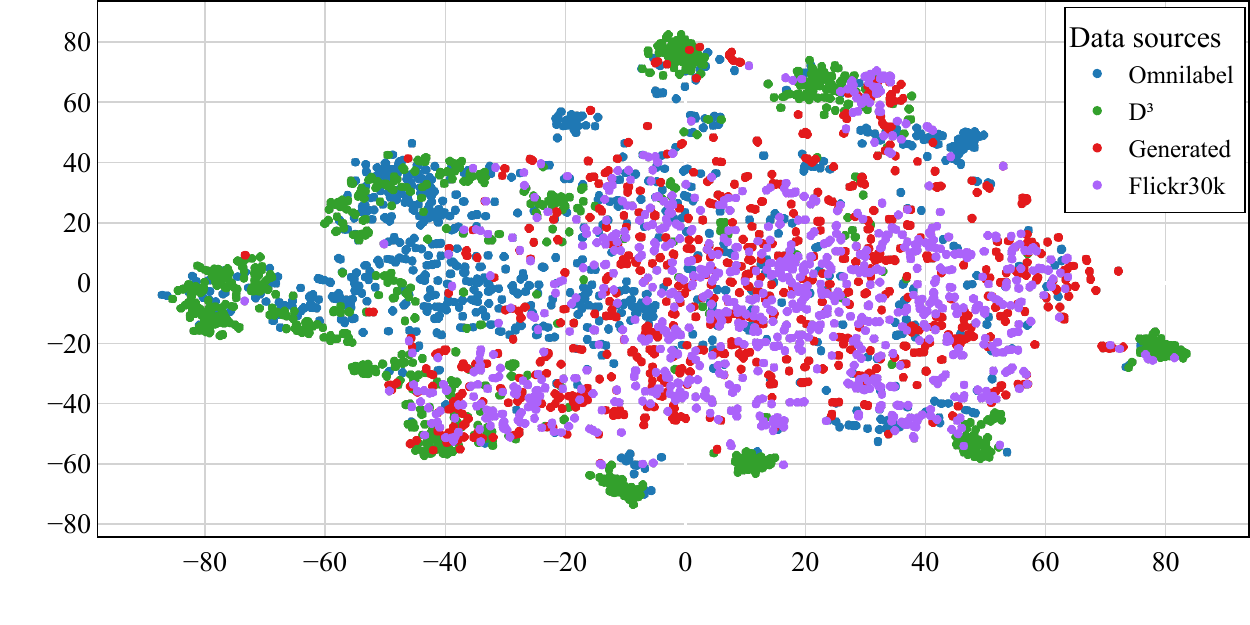}
  \vspace{-3mm}
  \caption{Distributions of image regions from Omnilabel, D$^3$, and our generated images. Visualization with t-SNE.}
  \label{fig:tSNE_negImg_diff_datasets}
\end{figure}

\paragraph{Looking into generated images and benchmarks:}
Table~\ref{tab:ablation_results_negimg_ol} shows that negative images help on OmniLabel but not much on D$^3$.
We explore this on a data basis.
We first crop image regions for generated images, OmniLabel images, D$^3$ images, and Flickr30k images based on the bounding boxes. 
Then, we randomly select 1000 image regions and feed them into a CLIP image encoder to get CLIP embeddings.
Later, we input those embeddings to t-SNE~\cite{vandermaaten_jmlr08_tsne} to illustrate the similarities between different image regions.
As shown in Fig.~\ref{fig:tSNE_negImg_diff_datasets}, D$^3$'s regions are grouped into several clusters, while OmniLabel and our generated regions are scattered in the center. 
This indicates that there is a clear domain gap between D$^3$ and the others. 
Thus, it is plausible that our generated images only helps on OmniLabel.
In our view, the gap comes from that D$^3$ collect data in groups based on categories. In contrast, OmniLabel collects data randomly.

%% file: tbls/tbl_main_results_ol.tex
\setlength{\tabcolsep}{4pt}
\renewcommand{\arraystretch}{1}
\begin{table}[t] 
\centering
{\footnotesize
\begin{tabular}{@{} l cccc cccc @{}}
\toprule
 & \multicolumn{4}{c}{\textbf{OmniLabel}}
 & \multicolumn{4}{c}{\textbf{OmniLabel-Negative}} 
 \\

\cmidrule(r){2-5} \cmidrule(lr){6-9} 

 & \textbf{AP} & \textbf{APc} & \textbf{APd} & \textbf{APdP}
 & \textbf{AP} & \textbf{APc} & \textbf{APd} & \textbf{APdP} 
 \\

\midrule

Detic~\cite{zhou2022detecting}
    & 8.0 & 15.6 & 5.4 & 8.0 
    & - & - & - & -
    \\
MDETR~\cite{kamath_iccv_21}
    & - & - & 4.7  & 9.1
    & - & - & - & -
    \\

\midrule

GLIP-T~\cite{li2022grounded}
    & 19.3 & 23.6 & 16.4 & 25.8
    & 13.9 & 24.8 & 9.6 & 26.1
    \\
+ Ours
    & 22.2 & 27.2 & 18.8 & 29.0 
    & 16.5 & 28.6 & 11.6 & 30.2
    \\
\midrule
FIBER-B~\cite{dou2022coarse_fiber}
    & 25.7 & 30.3 & 22.3 & 34.8
    & 18.7 & 31.2 & 13.3 & 36.3
    \\
+ Ours
    & 28.1 & 32.1 & 25.1 & 36.5
    & 22.3 & 33.3 & 16.7 & 38.3
    \\

\bottomrule
\end{tabular}
}
\caption{Evaluation on the OmniLabel~\cite{schulter2023omnilabel} benchmark.}
\label{tab:main_results_ol}
\end{table}
\setlength{\tabcolsep}{1.4pt}
\renewcommand{\arraystretch}{1}

%% file: tbls/tbl_main_results_d3.tex
\setlength{\tabcolsep}{5.5pt}
\renewcommand{\arraystretch}{1}
\begin{table}[t] 
\centering
{\footnotesize
\begin{tabular}{@{} l ccc cccc @{}}
\toprule
 & \multicolumn{3}{c}{\textbf{D$^3$ (default)}}
 & \multicolumn{4}{c}{\textbf{D$^3$ (by length of texts)}}
 \\

\cmidrule(r){2-4} \cmidrule(lr){5-8}

 & \textbf{Full} & \textbf{Pres} & \textbf{Abs}
 & \textbf{S} & \textbf{M} & \textbf{L} & \textbf{XL} 
 \\

\midrule

OFA-L~\cite{wang_icml22_ofa}
    & 4.2 & 4.1 & 4.6
    & 4.9 & 5.4 & 3.0 & 2.1 \\
OWL-ViT-L~\cite{minderer_eccv22_owlvit}
    & 9.6 & 10.7 & 6.4 
    & 20.7 & 9.4 & 6.0 & 5.3 \\
G-DINO-B~\cite{liu2023_gdino}
    & 20.7 & 20.1 & 22.5 
    & 22.6 & 22.5 & 18.9 & 16.5 \\
OFA-DOD~\cite{xie2023dcube}
    & 21.6 & 23.7 & 15.4
    & 23.6 & 22.6 & 20.5 & 18.4 \\

\midrule

GLIP-T~\cite{li2022grounded}
    & 19.1 & 18.3 & 21.5
    & 22.4 & 22.0 & 16.6 & 10.6 \\
+ Ours
    & 21.4 & 20.6 & 23.7
    & 28.1 & 24.5 & 17.4 & 11.5 \\
    
\midrule
FIBER-B~\cite{dou2022coarse_fiber}
    & 22.7 & 21.5 & 26.0
    & 30.1 & 25.9 & 17.9 & 13.1 \\
+ Ours
    & 26.0 & 25.2 & 28.1
    & 35.5 & 29.7 & 20.5 & 14.2 \\

\bottomrule
\end{tabular}
}
\caption{Evaluation on the D$^3$~\cite{xie2023dcube} benchmarks.}
\label{tab:main_results_d3}
\end{table}
\setlength{\tabcolsep}{1.4pt}
\renewcommand{\arraystretch}{1}

%% file: sec/5_conclusion.tex
\section{Conclusion}
\label{sec:conclusion}

Language-based detection requires localization of objects by a referring free-form text descriptions. To train accurate models in a discriminative way, the training data must contain good negative samples. Starting with an existing dataset, we propose (1) novel ways to prompt LLMs for generating additional negative texts, and (2) generating negative images to complement the training signal. Based on our experimental evaluations, we conclude that such additional negative training data indeed translates into improved detection accuracy on standard benchmarks. Our analysis demonstrates the importance of diversity in the generated text, which is higher with our approach than with prior works, and the quality of the generated images, which our proposed filtering steps can significantly increase.

\noindent {\bf Acknowledgments:} 
This research project has been partially funded by research grants to Dimitris N. Metaxas through NSF: 2310966, 2235405, 2212301, 2003874, and FA9550-23-1-0417.

%% file: sec/X_suppl.tex
\clearpage
\setcounter{page}{1}
\maketitlesupplementary

\noindent
This document supplements the main paper as follows.
\begin{itemize}
  \item Sect.~\ref{sect:supp_rule-based_negTxt} describes the generation of rule-based negative texts.
  \item Sect.~\ref{sect:supp_prompts_negTxt} provides prompts to LLMs for negative text generations.
  \item Sect.~\ref{sect:supp_combine_negTxt} explores the combinations of negative texts generated by different methods.
  \item Sect.~\ref{sect:supp_subject_study_negTxt} provides the subject studies on negative texts.
  \item Sect.~\ref{sect:supp_vis_negTxt} visualizes good and failure cases of negative texts.
  \item Sect.~\ref{sect:supp_negImg} provides more details about negative image generation, including prompts to LLMs.
  \item Sect.~\ref{sect:supp_vis_negImg} visualizes good and failure cases of negative images.
  \item Sect.~\ref{sect:supp_more_negImg_filter} elaborates Box and CLIP filters for negative images.
  \item Sect.~\ref{sect:supp_concat_negImg} illustrates the concatenation of positive and negative images for finetuning with negative images.
  \item Sect.~\ref{sect:supp_impl_detail} provides implementation details.
\end{itemize}

\begin{lstfloat}[b]
{\footnotesize
\begin{lstlisting}[caption={Prompt to ChatGPT for extracting concepts (i.e., objects, attributes, and relationships) from a text.}, captionpos=b, breaklines=true, label=lst:prompt_extract_concepts]
Find objects, attributes of objects, and relationships between objects in the given text. One word or phrase should only be one of objects, attributes and relationships.

Provide the answer using the JSON format as,
{
"text": <the given text>,
"objects": <list of attributes in the given text>
"attributes": <list of attributes in the given text>
"relationships": <list of attributes in the given text>
}

Given text: {positive_text}
\end{lstlisting}
}
\end{lstfloat}

\begin{lstfloat}[b]
{\footnotesize
\begin{lstlisting}[caption={Prompt to ChatGPT for generating negative texts by changing one concept.}, captionpos=b, breaklines=true, label=lst:prompt_foil_one_concept]
Given the positive text that refers to objects in an image, your task is to generate hard negative texts that do not refer to any objects in the image. A list of phrases in the positive text will be given. You need to replace each of phrases with other words to generate hard negative texts.

Instructions:
1. Choice one phrase from the given phrase list in a sequence.
2. Find similar but different alternative concepts for the picked phrase.
3. Choice one alternative to replace the phrase to generate a negative text. Keep other words intact as much as possible. Maintain the structure of the sentence. The resulting text should be coherent and satisfy common senses.
4. For each phrase, generate at least 2 negative texts.
5. Please provide the results in a JSON format as 
{
"positive_text": <positive text>, 
"results": [
{
"phrase": <phrase >
"negative_texts": [list of negatives texts by changing the phrase]
}, ......
{
"phrase": <phrase >
"negative_texts": [list of negatives texts by changing the phrase]
},]}

The input text: {input_text}
Phrase list: {list_of_phrases_in_input_text}
\end{lstlisting}
}
\end{lstfloat}

\begin{lstfloat*}[t]
{\footnotesize
\begin{lstlisting}[caption={Examples of instruction-output pairs. A LLaMA-7B model is finetuned with such pairs to foil concepts.}, captionpos=b, breaklines=true, label=lst:llama_foil_one_example]
[{
"instruction": "Change the given phrase and output a negative text that is semantically different from the positive text \"Rows of adults taking in a lecture in a classroom.\". Given phrase: adults", 
"output": "Groups of children taking in a lecture in a classroom."
},
{
"instruction": "Change the given phrase and output a negative text that is semantically different from the positive text \"Rows of adults taking in a lecture in a classroom.\". Given phrase: lecture", 
"output": "Rows of adults participating in a discussion in a classroom."
}, 
...]
\end{lstlisting}
}
\end{lstfloat*}

\section{More about negative texts}

\subsection{Rule-based negative text generation} \label{sect:supp_rule-based_negTxt}
We extract concepts (i.e., objects, attributes and relationships) from the original caption using the Scene Graph Parser\footnote{\url{https://github.com/vacancy/SceneGraphParse}}.  
Then, we generate a negative caption by changing or foiling one concept in the caption.  Taking as an example the caption ``A boy is playing with a dog'', we can change ``A boy'' into ``A girl'', and get the negative caption ``A girl is playing with a dog''.  Such foils require candidate concepts that do not refer to the same object as the original concept. 
To get good candidates, we follow~\cite{ma_cvpr23_crepe} to use WordNet, as well as other external knowledge bases, i.e., ConceptAPI, and human annotations~\cite{bogin2021covr}.

\begin{lstfloat*}[t]
{\footnotesize
\begin{lstlisting}[caption={Prompt to ChatGPT for generating negative texts by recombining objects of the original positive caption.}, captionpos=b, breaklines=true, label=lst:prompt_recombination]
You are asked to generate 10 sentences using a list of text phrases extracting from a main sentence.

Here are the requirements:
1. Be confident that the generated sentences should be semantically different from the main sentence.
2. You need to keep at least one phrase intact
3. You can either change one phrase to a closed but different concepts, e.g. "man" into "woman", "dog" into "cat". Or you can add new text phrases not in the list
4. Focus on new relationships that are not in the main sentence
5. Avoid general descriptive sentences because they may refer to the same object as the main sentence, e.g. "book is helpful"
6. Keep the generated sentences simple

The results should be in JSON format as
{
"main": <main sentence>,
"generated": [<generated sentence>, ......, <generated sentence>]
}

Main sentence: {positive_text}
Text phrase list: {phrase_list}
\end{lstlisting}
}
\end{lstfloat*}

\subsection{Prompts to LLMs for negative text generation}
\label{sect:supp_prompts_negTxt}

As mentioned in the main paper, we first leverage a strong instruction-tuned LLM~\cite{openai2022gpt-3.5} to generate 50k positive-negative pairs, and then finetune a LLaMA-7B~\cite{touvron2023llama} model with those pairs to later generate negative captions on large grounding datasets. 
We describe prompts we used to generate positive-negative pairs for finetuning as follows.

\paragraph{Prompts for LLM-based foils:} 
We follow a ChatGPT-LLaMA strategy to generate negative captions at scale. 
That is, first, we use ChatGPT to extract concepts, i.e., objects, attributes, and relationships, from some positive captions that are randomly selected from the visual grounding dataset Flickr30k~\cite{plummer2015flickr30k}. The detailed prompt for this step is provided in Listing~\ref{lst:prompt_extract_concepts}.
Then, we ask ChatGPT to change or foil each concept in the positive caption sequentially to get negative captions. Listing~\ref{lst:prompt_foil_one_concept} presents the detailed prompt. In this step, we generate around 50k positive-negative pairs.
Finally, a LLaMA-7B model is instruction-finetuned with the instruction-output pairs that are built upon the generated positive-negative pairs. Listing~\ref{lst:llama_foil_one_example} shows exemplar pairs.
We use the same instruction as finetuning during inference to get outputs for all captions of Flickr30k.

\paragraph{Prompts for re-combination with LLMs:}
We follow a similar ChatGPT-LLaMA strategy as LLM-based foils.
First, we use the same prompt (See Listing~\ref{lst:prompt_extract_concepts} again) as LLM-based foils to get concepts from randomly selected positive captions of Flickr30k.
Then, we ask ChatGPT to re-combine all objects of each caption to get new negative captions. Listing~\ref{lst:prompt_recombination} presents the detailed prompt. 
Finally, we also finetune a LLaMA-7B model but with slightly different instruction-output pairs as exhibited in Listing~\ref{lst:llama_recombine_example}.

\paragraph{Prompts for in-context summary with LLMs:}
We first ask ChatGPT to make a summary of positive-negative pairs from Winoground~\cite{thrush_cvpr2022_winoground} using the prompt shown in Listing~\ref{lst:prompt_summary}.
Then, as shown in Listing~\ref{lst:prompt_incontext_summary_negTxt}, the summary is used as part of the prompt of ChatGPT to generate more positive-negative pairs, with Winoground pairs as in-context examples.
Finally, based on those positive-negative pairs, we build instruction-output pairs (See examples in Listing~\ref{lst:llama_incontext_summary_example}) to finetune a LLaMA-7B model.

\begin{lstfloat}[t]
{\footnotesize
\begin{lstlisting}[caption={Examples of instruction-output pairs. A LLaMA-7B model is finetuned with such pairs to recombine objects and create a negative text.}, captionpos=b, breaklines=true, label=lst:llama_recombine_example]
[{
"instruction": "Use the input phrase list to generate 10 texts that are semantically different from the positive text \"A man wearing a green striped shirt while jumping up onto a mountain.\". Phrase list: man, shirt, mountain",
"output": "[\"A woman wearing a blue polka dot shirt while climbing down from a mountain.\", \"A man wearing a red plaid shirt while running along a beach.\", \"A man wearing a yellow striped shirt while hiking through a forest.\", \"A man wearing a green striped shirt while skiing down a snowy slope.\", \"A man wearing a green striped shirt while riding a bike on a trail.\", \"A man wearing a green striped shirt while swimming in a lake.\", \"A man wearing a green striped shirt while exploring a cave.\", \"A man wearing a green striped shirt while fishing in a river.\", \"A man wearing a green striped shirt while camping in the woods.\", \"A man wearing a green striped shirt while paragliding from a cliff.\"]"
}, ...]
\end{lstlisting}
}
\end{lstfloat}

\begin{lstfloat}[t]
{\footnotesize
\begin{lstlisting}[caption={Prompt to ChatGPT for making a summary of 80 positive-negative pairs from Winoground. Only the first 4 pairs are displayed.}, captionpos=b, breaklines=true, label=lst:prompt_summary]
Check the following pairs of the input text and its negative text.

###
Input text: an old person kisses a young person
hard negative: a young person kisses an old person
###
Input text: a person with short hair is happily ironing a light blue shirt while a person with long hair sits on a white couch
hard negative: a person with long hair is happily ironing a light blue shirt while a person with short hair sits on a white couch
###
Input text: the masked wrestler hits the unmasked wrestler
hard negative: the unmasked wrestler hits the masked wrestler
###
Input text: a person watches an animal
hard negative: an animal watches a person
###
[...]

Summarize the features of those pairs.
\end{lstlisting}
}
\end{lstfloat}

\begin{figure}[tb]
  \centering
  \includegraphics[width=1.0\linewidth]{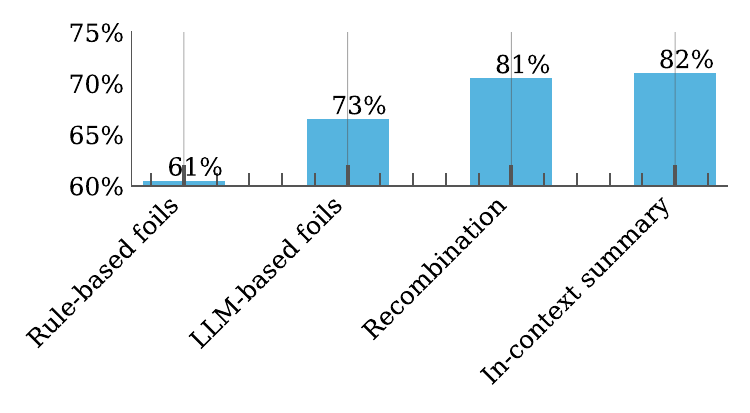}
  \caption{Percentage of good generated negative texts for 100 randomly selected samples.}
  \label{fig:supp_negTxt_subject_study}
\end{figure}

\begin{lstfloat*}[t]
{\footnotesize
\begin{lstlisting}[caption={Prompt to ChatGPT for generating negative texts with a summary and in-context samples. 
The summary is generated by ChatGPT and provides contexts about positive-negative pairs. The first three pairs are in-context samples randomly selected from Winoground.
We use the completion API to get more pairs beyond in-context samples.}, captionpos=b, breaklines=true, label=lst:prompt_incontext_summary_negTxt]
The pairs of input text and hard negative text share the following features:
1. The overall structure and elements of the scene remain the same in both the input text and the hard negative text.
2. The main action or relationship portrayed in the input text is reversed or altered in the hard negative text.
3. In some cases, the positions or attributes of the objects or characters in the scene are swapped or modified.
4. The language and syntax used in the input text are generally maintained, but the order or arrangement of the words may be adjusted in the hard negative text to create a contrasting effect.
5. The hard negative text often introduces a contradictory or unexpected element that contradicts the information in the input text.
6. The hard negative text may highlight a different perspective or focus on a different aspect of the scene compared to the input text.
7. In some cases, the hard negative text uses antonyms or contrasting adjectives to create a contrasting effect.
8. The hard negative text may play with the order of events in the scene, causing a shift in the temporal relationship between actions or elements.
9. The positioning or placement of objects or characters in relation to each other may be reversed or altered in the hard negative text.
10. The hard negative text may introduce additional details or new elements that create a contrasting or unexpected effect.

Generate 20 pairs of input and hard negative.
###
1. Input: an old person kisses a young person
1. Negative: a young person kisses an old person
###
2. Input: a person with short hair is happily ironing a light blue shirt while a person with long hair sits on a white couch
2. Negative: a person with long hair is happily ironing a light blue shirt while a person with short hair sits on a white couch
###
3. Input: the masked wrestler hits the unmasked wrestler
3. Negative: the unmasked wrestler hits the masked wrestler
###
4. Input:
\end{lstlisting}
}
\end{lstfloat*}

\begin{lstfloat}[t]
{\footnotesize
\begin{lstlisting}[caption={Examples of instruction-output pairs from in-context summary for finetuning LLaMA-7B to generate negative texts.}, captionpos=b, breaklines=true, label=lst:llama_incontext_summary_example]
[{
"instruction": "Generate 1 negative text that are semantically different from the positive text \"the cake is bigger than the plate it's on\"", "input": "", "output": "the plate is bigger than the cake that's on it"},
{"instruction": "Generate 1 negative text that are semantically different from the positive text \"the dog is chasing the ball and the owner is running after it\"", "input": "", "output": "the owner is chasing the ball and the dog is running after it"},
...]
\end{lstlisting}
}
\end{lstfloat}

\begin{lstfloat}[t]
{\footnotesize
\begin{lstlisting}[caption={Examples of instruction-output pairs from editing texts with controls}, captionpos=b, breaklines=true, label=lst:llama_edit_w_control]
[{
"instruction": "Generate 1 sentence by replacing [Mask] in \"A group of [Mask] and play on the playground.\". The generated sentence should be different from \"A group of children laugh and play on the playground.\"", 
"output": "A group of adults laugh and play on the playground."
}, 
{
"instruction": "Generate 1 sentence by replacing [Mask] in \"The [Mask] purrs contentedly on the windowsill.\". The generated sentence should be different from \"The cat purrs contentedly on the windowsill.\"", 
"output": "The dog purrs contentedly on the windowsill."
}
...]
\end{lstlisting}
}
\end{lstfloat}

\begin{lstfloat}[t]
{\footnotesize
\begin{lstlisting}[caption={Prompt to ChatGPT for generating text-mask-negative triplets with the completion API. The summary is generated by ChatGPT and provides contexts about the generation. The in-context samples are generated by ChatGPT with this prompt and human checks.}, captionpos=b, breaklines=true, label=lst:prompt_neg_text_w_control]
The example of a input and a output text share the following features:
- The input text demonstrate a wide range of scenarios and actions.
- One concept (that replaces a specific noun or attribute in the original sentence) in the input is masked with [Mask].
- The output replace [Mask] with a different noun or attribute that is related to the original sentence but provides a new context or perspective.
- The texts except the masked text remain the same in the output text.
- The replacement often involve changing the subject, object, or location of the sentence.
- Some examples involve changing attributes, e.g. colors, shape, or materials.
- The replacement sometimes involve changing the age or gender of characters.
- There are also examples where the substitutions involve changing objects or activities.
- The overall purpose of the output is to create a new sentence that has similar format as the original sentence but with a twist in meaning.

Generate 20 such examples
###
1. Input: Commuters wait for to cross a street.
1. Masked input: Commuters wait for to cross [Mask].
1. Output: Commuters wait for to cross a river.
###
2. Input: A child playing with a toy car in the park.
2. Masked input: A child playing with [Mask] in the park.
2. Output: A child playing with a soccer ball in the park.
###
3. Input: The squirrel is climbing up the tree.
3. Masked input: [Mask] is climbing up the tree.
3. Output: The monkey is climbing up the tree.
###
4. Input:
\end{lstlisting}
}
\end{lstfloat}

\subsection{Future analysis on negative texts} \label{sect:supp_more_analysis_negTxt}

\subsubsection{Combining negative texts of different sources} \label{sect:supp_combine_negTxt}
In this part, we explore to combine negative texts of different sources.
When two or more sources are combined, we randomly sample negative texts from the combined source during training.
Table~\ref{tab:supp_merge_neg_texts} provides the results of different combinations, from which we have the following findings.
First, all combinations outperform the original FIBER in terms of almost all the metrics, which indicates the effectiveness of adding negative texts.
Second, there is no combination that consistently outperforms others in all metrics.
For example, ``+ (4) Texts for negative images'' achieves the leading performance on the whole OmniLabel but not on OmniLabel-Negative and D$^3$.
Moreover, the performance gaps between different combinations are minor.
Some possible explanations to such results are: 
1) All four methods generate negative texts mainly by changing concepts of the original captions. Thus, negative texts of different methods are similar and not complementary.
2) It is most important to break the bias that all input texts are positive.
Adding more or less negative texts should act similarly in terms of breaking the bias.

\subsubsection{Subject studies on negative texts} \label{sect:supp_subject_study_negTxt}
We employ human experts to check the amount of low-quality negative texts that either have the same meaning as the original positive captions or are uncommon in the real world.
Please refer to  Fig.~\ref{fig:supp_bad_case_negTxt} for some low-quality examples.
We randomly selected 100 positive-negative text pairs with the corresponding images for each of rule-based foils, LLM-based foils, recombination, and in-context summary.
Two exports were asked to check one pair, and the pair was regarded as good when two of them agreed.
Fig.~\ref{fig:supp_negTxt_subject_study} shows the percentage of good cases for different methods. 
As shown, rule-based foils contain the most low-quality data. Recombination and in-context summary have a similar amount of good cases.
Please check Sect.~\ref{sect:supp_vis_negTxt} for more analysis and visualizations.

\begin{table*}
\centering
 \resizebox{.95\textwidth}{!}{ 
 \begin{tabular}{@{} l cccc cccc ccc cccc @{}}
 \toprule
  & \multicolumn{4}{c}{\textbf{Whole OmniLabel}} & \multicolumn{4}{c}{\textbf{OmniLabel-Negative}} & \multicolumn{3}{c}{\textbf{D$^3$}} (default) & \multicolumn{4}{c}{\textbf{D$^3$} (by length of texts)}
 \\ 
 \cmidrule(r){2-5} \cmidrule(lr){6-9} \cmidrule(lr){10-12} \cmidrule(lr){13-16}  

  & \textbf{AP} & \textbf{APc} & \textbf{APd} & \textbf{APd-P}
  & \textbf{AP} & \textbf{APc} & \textbf{APd} & \textbf{APd-P}
  & \textbf{Full} & \textbf{Pres} & \textbf{Abs}
  & \textbf{S} & \textbf{M} & \textbf{L} & \textbf{XL}
  \\ 
 \midrule

 Original FIBER-B
     & 25.7 & 30.3 & 22.3 & 34.8
     & 18.7 & 31.2 & 13.3 & 36.3
     & 22.7 & 21.5 & 26.0 
     & 30.1 & 25.9 & 17.9 & 13.1
      \\
 + (1) LLM-based foils
     & 26.5 & 30.7 & 23.3 & \underline{35.9}
     & 20.8 & 32.1 & 15.4 & \textbf{38.0} 
     & 24.6 & 24.0 & 26.5
     & 33.8 & 28.5 & 18.8 & 13.2
     \\
 + (2) Re-combination
     & {26.9} & \underline{30.8} & {23.9} & \underline{35.9}
     & \underline{21.1} & \underline{32.3} & {15.6} & {37.6} 
     & {25.3} & {24.6} & {27.3}
     & 33.6 & 29.1 & 19.8 & 14.3
     \\
 + (3) In-context summary
     & {26.6} & \underline{30.8} & {23.4} & 34.2
     & \underline{21.1} & {32.2} & \underline{15.7} & 36.4 
     & {25.7} & \underline{25.2} & {27.5}
     & \textbf{34.9} & 28.9 & 20.6 & \underline{15.0}
     \\ 
  + (4) Texts for negative images
    & \textbf{27.4} & \textbf{31.4} & \textbf{24.3} & \textbf{37.3}
    & 20.4 & \textbf{32.6} & 14.9 & \textbf{38.0}
    & 25.4 & 24.7 & 27.7
    & \underline{34.8} & 28.7 & \underline{20.8} & 13.0
     \\ 
  + (3)(4)
    & 26.9 & 30.7 & 23.9 & 35.5
    & \underline{21.1} & 32.1 & 15.6 & 37.0
    & \underline{25.9} & 25.1 & \textbf{28.5}
    & 34.4 & \textbf{30.0} & 20.1 & \underline{15.0}
     \\ 
  + (2)(3)(4)
    & \underline{27.0} & 30.7 & \underline{24.0} & {35.6}
    & 20.9 & 32.2 & 15.4 & 37.2
    & \textbf{26.1} & \textbf{25.3} & \underline{28.2}
    & 34.1 & \textbf{30.0} & \textbf{20.9} & \textbf{15.1}
     \\ 
  + (1)(2)(3)(4)
    & 26.9 & 30.7 & 23.9 & 35.3
    & \textbf{21.2} & 32.1 & \textbf{15.8} & 37.0
    & 25.2 & 24.4 & 27.6 
    & 34.1 & 29.0 & 19.5 & 14.0
     \\ 
 \bottomrule
 \end{tabular}
 }
  \caption{Performance of FIBER-B trained with different combinations of negative texts.}
 \label{tab:supp_merge_neg_texts}
\end{table*}

\subsubsection{Visualizations of negative texts} \label{sect:supp_vis_negTxt}

\begin{figure*}[tb]
  \centering
  \includegraphics[width=1.0\linewidth]{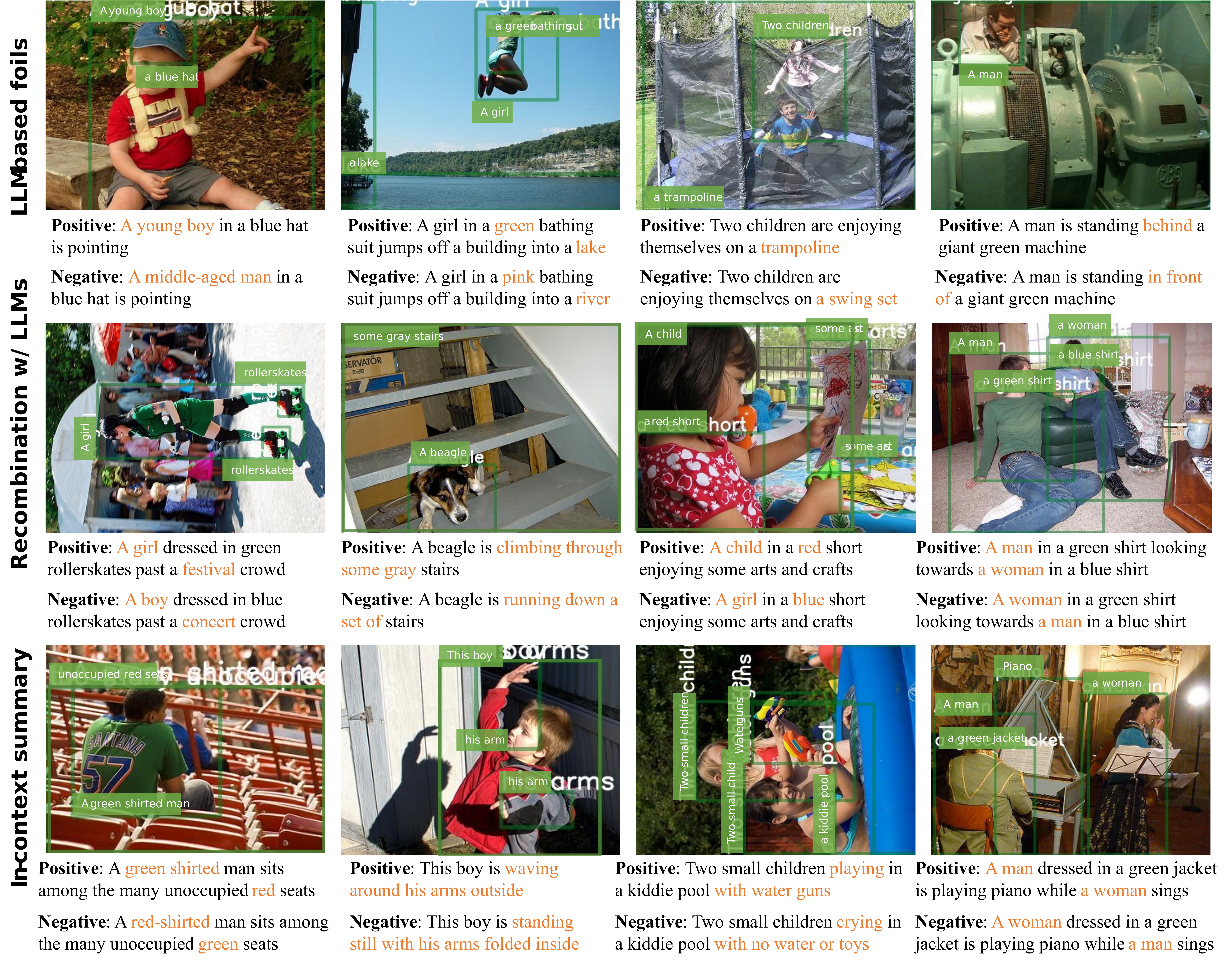}
  \caption{Good cases of negative texts from LLM-based foils, recombination, and in-context summary. The negative texts come with the positive texts and the corresponding images. We highlight text phrases before and after changes in orange, and objects referred by the original caption in green boxes.
  LLMs can replace objects, attributes, and relationships with similar but different concepts. Recombination and in-context summary involve the swap and negations of concepts.}
  \label{fig:supp_good_case_negTxt}
\end{figure*}

\paragraph{Visualizations of good cases:}
Fig.~\ref{fig:supp_good_case_negTxt} provides good cases of negative texts from three methods that adopt LLMs.
As shown, LLMs can replace objects, attributes, or relationships with other reasonable concepts.
Note that negative texts from our in-context summary include the swap of concepts, where no words were removed or added.
Since such swap is a feature of Winoground~\cite{thrush_cvpr2022_winoground}, we can conclude that the in-context summary is an effective way to learn specific patterns of a given dataset without the need of human curated prompts.

\begin{figure*}[tb]
  \centering
  \includegraphics[width=1.0\linewidth]{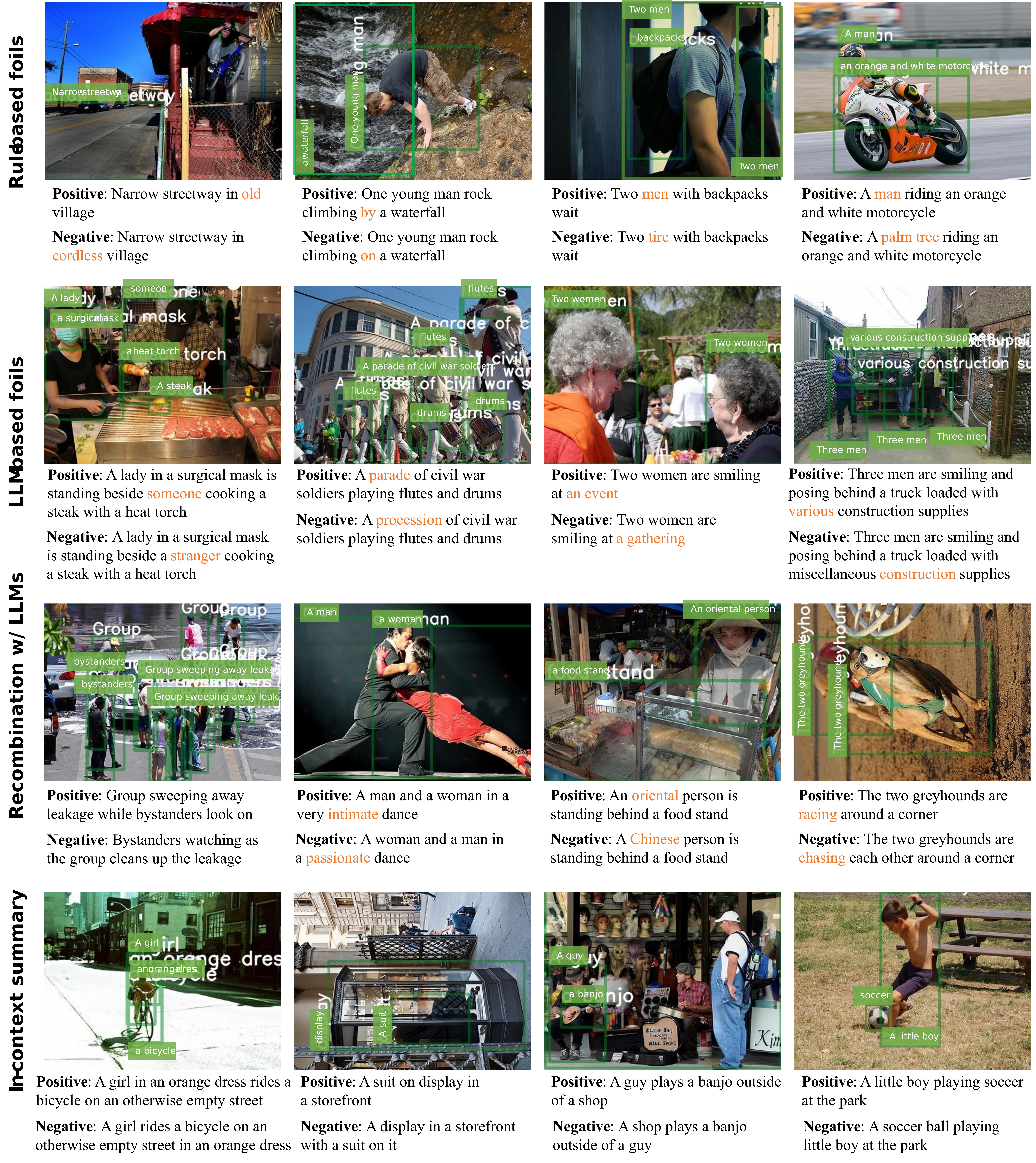}
  \caption{Failure cases of negative texts from four methods. 
  Rule-based foils generate uncommon texts that do not meet common sense. 
  LLM-based foils and recombination fail to replace concepts with discriminative concepts.
  In-context summary either does not change the meaning or generates uncommon texts by swap.
  \\
  \\
  \\
  }
  \label{fig:supp_bad_case_negTxt}
\end{figure*}

\paragraph{Visualizations of failure cases:}
As shown in Fig.~\ref{fig:supp_bad_case_negTxt}, the most common failure cases of rule-based foils are that the generated negative texts do not meet the common sense or logic. This is because rules just replace concepts with different concepts and do not consider the context of the original caption.
LLM-based foils and recombination may fail to replace original concepts with discriminative concepts so that the generated captions have the same meaning as the original captions.
In-context summary usually tries to swap concepts within the caption. But such swap does not always work in at least two aspects. 1) It does not change the meaning of the original caption. 2) It generates uncommon captions that do not meet the common sense or logic.

\begin{figure}[tb]
  \centering
  \includegraphics[width=1.0\linewidth]{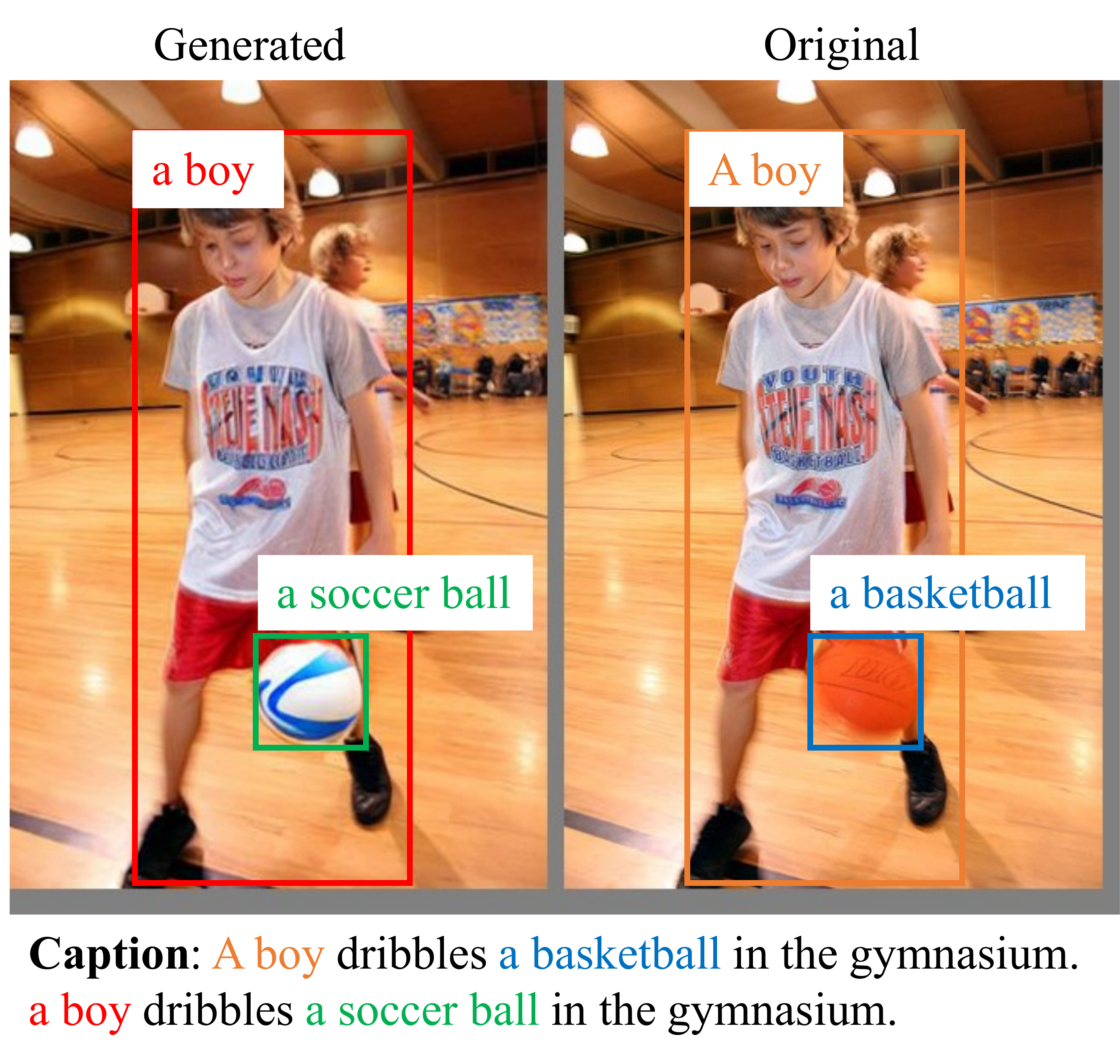}
  \caption{Concatenation of the positive image and corresponding generated negative image with the concatenation of captions.}
  \label{fig:supp_concat_img}
\end{figure}

\section{More about negative images}

\subsection{Negative image generation} \label{sect:supp_negImg}

\paragraph{Prompts for editing texts:}
We control which part of a text ChatGPT edits by masking out a phrase with ``[Mask]'' and ask it to get a negative text by filling the ``[Mask]'' without changing other part of the original text. 
As described in the main paper, we leverage our in-context summary to generate a number of text-mask-negative triplets.
We use the same prompt as Listing~\ref{lst:prompt_summary} except we replace Winoground pairs with text-mask-negative triplets.
Listing~\ref{lst:prompt_neg_text_w_control} shows the prompt for generating triplets with the summary and in-context samples.
Based on those triplets, we build instruction-output pairs to finetune a LLaMA-7B model. See Listing~\ref{lst:llama_edit_w_control} for an example of instruction-output pairs. Then, we ``[Mask]'' phrases in captions of Flickr30k, which correspond to some bounding boxes, and leverage the finetuned LLaMA model to edit texts with controls.

\paragraph{More details about conditional image generation:}
With the above edited negative texts with controls, the alignment between bounding boxes and edited phrases are remained. We then adopt GLIGEN~\cite{li2023gligen} to edit the content of bounding boxes based on the corresponding edited phrases.
GLIGEN is a finetuned variant of the stable diffusion model~\cite{rombach2022high}, which takes bounding boxes and descriptions of those boxes as conditions for image generation.
Refer to ~\cite{li2023gligen} for more technical details on the image generation.

\subsection{Visualizations of negative images} \label{sect:supp_vis_negImg}

\begin{figure*}[tb]
  \centering
  \includegraphics[width=1.0\linewidth]{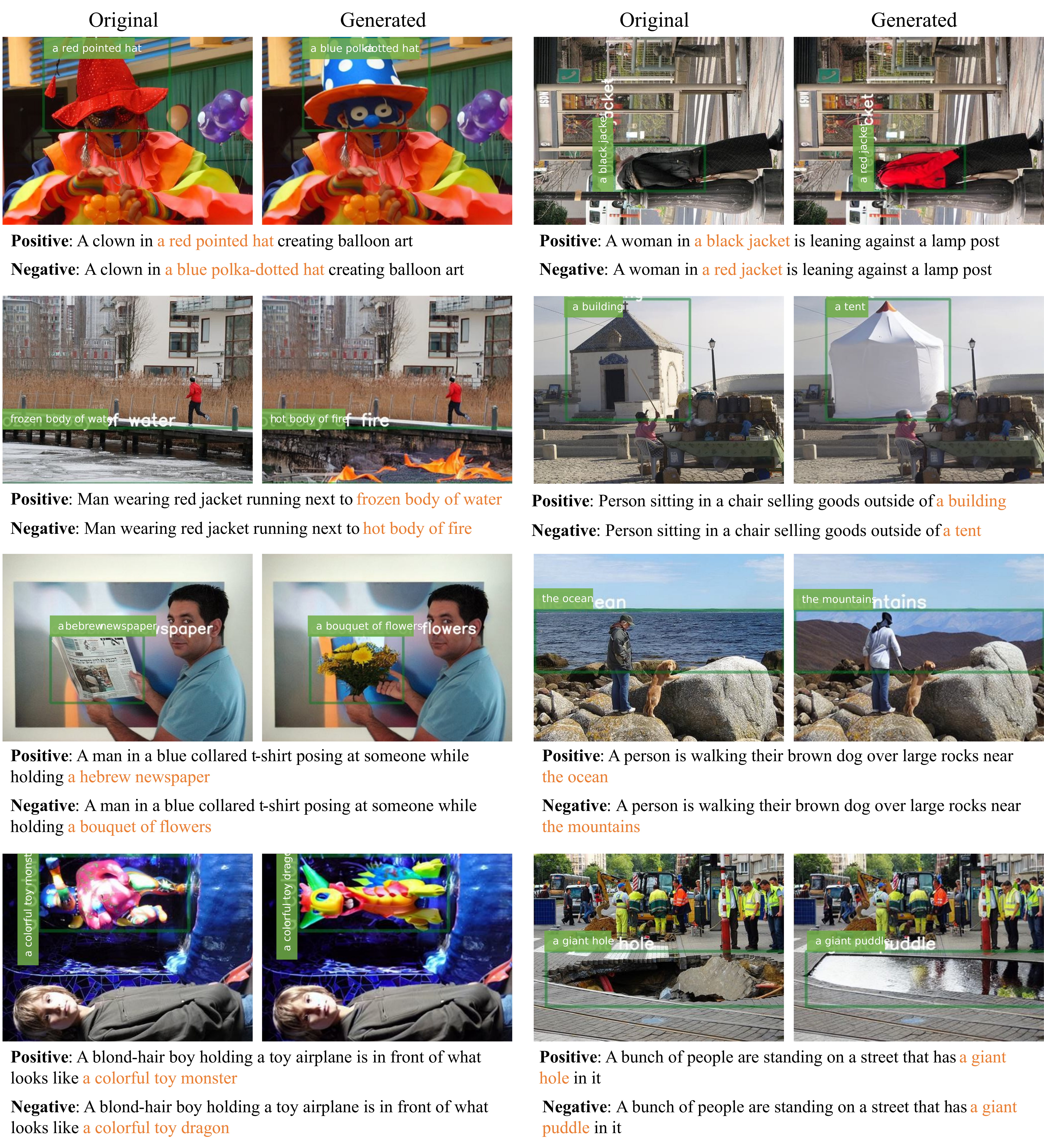}
  \caption{Good cases of generated negative images.}
  \label{fig:supp_good_case_negImg}
\end{figure*}

\paragraph{Visualizations of good cases:}
As shown in Fig.~\ref{fig:supp_good_case_negImg}, our negative image generation pipeline can provide a variety of negative images, including changing foreground objects, background, and attributes.

\begin{figure*}[tb]
  \centering
  \includegraphics[width=1.0\linewidth]{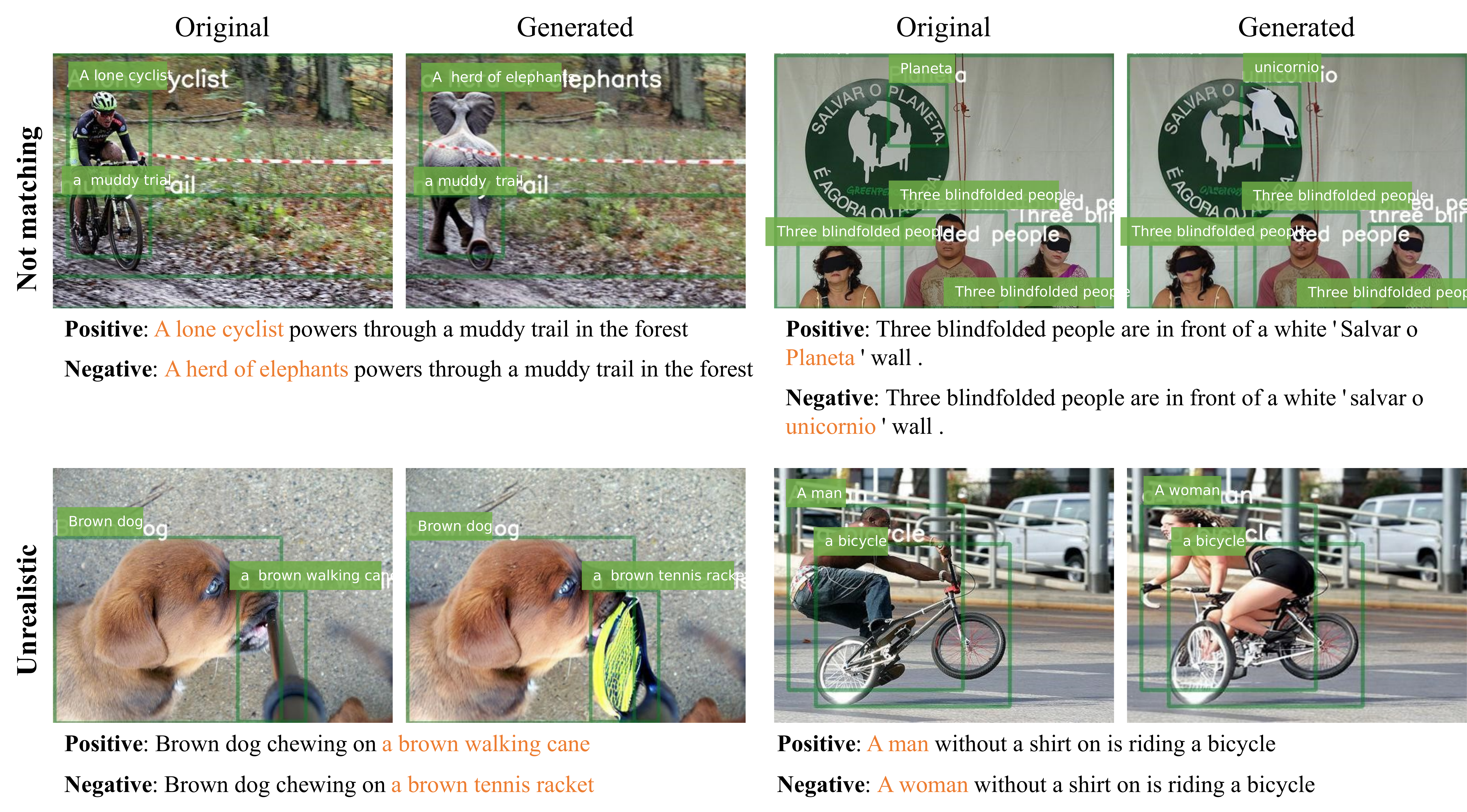}
  \caption{Two major types of failure cases for generated negative images, i.e. generated contents 1) not matched with the text phrase, or 2) unrealistic. Top left: Only one elephant not a herd. Top right: Not a text. Bottom left: The racket is distorted and not brown. Bottom right: The woman has more than four legs, and the bicycle is distorted.}
  \label{fig:supp_bad_case_negImg}
\end{figure*}

\paragraph{Visualizations of failure cases:}
After Box and CLIP filters, our generated negative images can still be wrong or in low-quality.
Fig.~\ref{fig:supp_bad_case_negImg} illustrates two major types of failure cases.
That is, generated contents 1) do not exactly match the text phrase, or 2) are unrealistic. 
We believe that the first type of failures can be mitigated when more effective VLMs are available for our CLIP filtering step.
The second type of failures are mainly caused by the limited capability of current image generative models, which are expected to be solved by the development of generative models.

\subsection{More details about Box and CLIP Filters} \label{sect:supp_more_negImg_filter}

\paragraph{Box filter:}
We generate negative images based on each phrase from Flickr30k.
If any boxes associated to the phrase covers more than 75\% any other boxes in the image, we ignore this phrase and move to the next phrase. 

\paragraph{CLIP filter:}
The CLIP filter consists of a image-caption level filtering and a box-phrase level filtering.
For the image-caption level filtering, we first feed to CLIP the whole generated image (visual input) and both positive and negative texts (text input). In this step, we get the unnormalized similarity score (i.e., logit) between the image and the positive text, and the logit between the image and the negative text.
Then, we apply a softmax on the two logits to get the normalized similarity score between the image and the negative text.
If the score is lower than 0.35, we drop the generated image.
For the box-phrase level filtering, we crop image regions based on bounding boxes whose text phrases are changed. Following VL-PLM~\cite{zhao2022exploiting}, we upscale boxes by a factor of 1.5 to include some contexts.
Then, we take cropped image region as visual inputs, and the original and modified text phrases are text input to CLIP. 
With the same scoring method as the image-caption level filtering, we compute the normalized similarity score between a cropped image region and the corresponding modified text phrase. 
If any cropped image region of an image has a score lower than 0.75, we drop this image.
We adopt the biggest open-sourced CLIP model, ``ViT-bigG-14'', from OpenCLIP\footnote{\url{https://github.com/mlfoundations/open_clip}}. 
After the two level filtering, around 46 \% of generated images are removed.

\subsection{Concatenating positive and negative images} \label{sect:supp_concat_negImg}

Assuming that the original image $I$ and the negative image $I'$ are given. Captions $t$ and $t'$ are corresponding to $I$ and $I'$, respectively.
We first shuffle and concatenate $I$ and $I'$ along either width or height, which is longer, as the image input during training. Note that $I$ can be either to the right or the left of $I'$ due to the shuffling if concatenated along the height.
Then, we concatenate $t$ and $t'$ as the text input.
Finally, we adjust the alignment between bounding boxes (in both $I$ and $I'$) and text phrases (in both $t$ and $t'$).
Fig.~\ref{fig:supp_concat_img} illustrates this concatenation.
As shown, the original image is on the right of the concatenated image. The caption ``A boy dribbles a basketball in the gymnasium'' only associates with the orange and the blue boxes.
The generated image is on the left, and its caption ``a boy dribbles a soccer ball in the gymnasium.'' only associates with the red and the green boxes.
The red and the orange boxes become negative to each other within one image input, although they contain the same boy.

\section{Implementation details} \label{sect:supp_impl_detail}
Our experiments are conducted on a computer with 8 NVIDIA A6000 GPUs.
All LLaMA models used in this paper are instruction-finetuned with low-rank adaptation (alpaca-lora)\footnote{\url{https://github.com/tloen/alpaca-lora}}.
Default hyper-parameters of alpaca-lora are adopted.
The finetuning is efficient and takes around 3 hours on our computer.

For finetuning of detectors, we initialize GLIP or FIBER models with their official checkpoints. 
We finetune both GLIP and FIBER a batch size of 16 and a learning rate of 1e$^{-5}$ for 1 epoch. 
Other hyper-parameters are kept the same as the official training.
The finetuning takes around 14 hours for FIBER and 18 hours for GLIP.